\documentclass[final,3p,times]{article}

\usepackage{amssymb}
\usepackage{amsmath} 
\usepackage{PRIMEarxiv}

\usepackage{latexsym}
\usepackage[latin1]{inputenc}
\usepackage{times}
\usepackage[linesnumbered,ruled,lined,boxed]{algorithm2e}
\usepackage{algorithmic}

\usepackage{epsfig,shadow,amssymb,bm}
\usepackage{booktabs}
\usepackage{multirow}
\usepackage{graphicx}
\usepackage{tikz}
\usepackage{stackrel}
\usepackage{pdflscape}[2016/05/14]
\usepackage{rotating}
\usepackage{tablefootnote}
\usepackage{dsfont}
\usepackage{hyphenat}

\usepackage{longtable}
\usepackage{enumitem}

\usepackage{colortbl}
\usepackage{xurl}
\usepackage{subcaption}
\usepackage[breaklinks=true]{hyperref}

\usetikzlibrary{shapes,arrows}
\usetikzlibrary{positioning}
\usetikzlibrary{arrows,automata,decorations,shapes}
\tikzset{
	treenode/.style = {shape=rectangle, rounded corners,
		draw, align=center,
		top color=white, bottom color=blue!20},
	root/.style     = {treenode, font=\Large, bottom color=red!30},
	env/.style      = {treenode, font=\ttfamily\normalsize},
	dummy/.style    = {circle,draw}
}

\newcommand{\CASE}[1]{\STATE \textbf{case} #1\textbf{:} \begin{ALC@g}}
	\newcommand{\ENDCASE}{\end{ALC@g}}

\newcommand{\DEFAULT}{\STATE \textbf{default:} \begin{ALC@g}}
	\newcommand{\ENDDEFAULT}{\end{ALC@g}}
\newcommand{\DEFAULTLINE}[1]{\STATE \textbf{default:} }

\title{Permutation-based multi-objective evolutionary feature selection for high-dimensional data}

\author{Raquel Espinosa, , Gracia S\'anchez, Jos\'e Palma, Fernando Jim\'enez\\
	Department of Information and Communications Engineering \\
	Faculty of Informatics \\
	University of Murcia\\ 
	Spain \\
	\texttt{\{raquel.espinosa,gracia,jtpalma,fernan\}@um.es} \\
}

\begin{document}
	\maketitle
	\interfootnotelinepenalty=10000
	
	\begin{abstract}
Feature selection is a critical step in the analysis of high-dimensional data, where the number of features often vastly exceeds the number of samples. Effective feature selection not only improves model performance and interpretability but also reduces computational costs and mitigates the risk of overfitting. In this context, we propose a novel feature selection method for high-dimensional data, based on the well-known permutation feature importance approach, but extending it to evaluate subsets of attributes rather than individual features. This extension more effectively captures how interactions among features influence model performance. The proposed method employs a multi-objective evolutionary algorithm to search for candidate feature subsets, with the objectives of maximizing the degradation in model performance when the selected features are shuffled, and minimizing the cardinality of the feature subset. The effectiveness of our method has been validated on a set of 24 publicly available high-dimensional datasets for classification and regression tasks, and compared against 9 well-established feature selection methods designed for high-dimensional problems, including the conventional permutation feature importance method. The results demonstrate the ability of our approach in balancing accuracy and computational efficiency, providing a powerful tool for feature selection in complex, high-dimensional datasets.
		
	\end{abstract}
	
	\keywords{
		Permutation-based feature selection \and high-dimensional data \and  multi-objective evolutionary algorithms \and classification \and regression
	}
	
\section{Introduction} 
In the realm of high-dimensional data analysis, \textit{feature selection} (FS) \cite{10.5555/2843983} is a critical step for improving the performance, interpretability, and generalizability of machine learning models. High-dimensional datasets, characterized by a large number of attributes relative to the number of samples, pose significant challenges, including increased computational cost, the risk of overfitting, and difficulties in model interpretation. FS methods aim to address these challenges by identifying a subset of relevant features that contribute most to the predictive performance of the model.

FS methods can be broadly categorized into \textit{filter}, \textit{wrapper}, and \textit{embedded} methods \cite{kuhn2020feature}. Filter methods operate independently of any learning algorithm and typically evaluate features based on their intrinsic properties, such as correlation with the target variable. 
Wrapper methods, by contrast, evaluate features based on the performance of a specific learning algorithm. While wrapper methods are often more accurate than filter methods due to their direct optimization of model performance, they are also significantly more computationally expensive, particularly when dealing with high-dimensional data. This computational burden often renders wrapper methods impractical for large-scale datasets.
Embedded methods integrate FS directly into the model training process. They balance the trade-off between accuracy and computational efficiency better than wrappers but are limited by their dependency on specific learning algorithms, which may lead to biased or suboptimal feature selection.

FS methods can be further divided into \textit{attribute evaluation} methods, such as \textit{feature ranking} (FR) methods, which rank individual features, and \textit{subset evaluation} methods, which assess the quality of subsets of features and require a strategy to search for candidate subsets of attributes. Subset evaluation methods are inherently \textit{multivariate}, as they consider the interactions between features, whereas attribute evaluation methods can be either \textit{univariate} or multivariate, depending on whether they evaluate features in isolation or in the context of other features.

An alternative approach that has gained attention is \textit{permutation feature importance} (PFI), which assesses the importance of features based on their impact on model performance when their values are randomly shuffled.
This randomization disrupts the original relationship between the feature and the target variable, allowing us to measure the feature's contribution to the model's accuracy. PFI seeks to maximize the degradation in model performance when a feature's values are permuted. The underlying goal is to identify features that, when disrupted, cause the most significant degradation in model performance, thereby indicating their importance in making accurate predictions. PFI is considered a multivariate method because it evaluates the importance of each feature by measuring the change in model performance when the feature's values are permuted, taking into account the context of all other features in the dataset. This approach captures the interactions between features and their collective impact on the model's predictive accuracy.

In this paper, we propose a novel feature selection method that extends PFI by evaluating subsets of attributes rather than individual features. Our method utilizes a \textit{multi-objective evolutionary algorithm} (MOEA) to search for optimal feature subsets, with objectives that include maximizing the degradation in model performance when subset feature values are permuted, and minimizing the cardinality of the feature subsets.  The proposed permutation-based feature selection method, which evaluates subsets of attributes rather than individual ones, goes a step further in reflecting feature interactions. We call the proposed method PSEFS-MOEA (\textit{permutation-based subset evaluation feature selection with MOEA}). By considering the collective effect of attribute subsets, our method more effectively captures complex interdependencies among features, leading to a more accurate and holistic selection of relevant attributes.
Then this approach offers a significant advantage in high-dimensional settings, where the complexity of feature interactions can greatly impact model performance. At the same time, it is computationally more feasible than wrapper methods, making it particularly suitable for high-dimensional data.  By combining the strengths of multivariate subset evaluation with the efficiency of permutation-based importance measures, the proposed method provides a robust and scalable solution for FS in complex, high-dimensional datasets. The main contributions of this work can be summarized as follows:

\begin{enumerate}
	\item We propose PSEFS-MOEA, a novel permutation-based feature selection method that evaluates subsets of attributes instead of individual features. By exploring complex attribute interactions and dependencies, PSEFS-MOEA overcomes the limitations of conventional permutation feature importance methods and other single-feature evaluation approaches.

	\item Two complementary versions of the method are introduced: PSEFS-MOEA-V1, which evaluates attribute subsets on a validation set, and PSEFS-MOEA-V2, which performs evaluations directly on the training set. 

	\item The method is tested on 24 datasets, encompassing a variety of tasks (binary, multi-class, imbalanced classification, and regression), ensuring robust evaluation across real-world scenarios. The implementation is straightforward and applicable to all task types without modification.

	\item PSEFS-MOEA achieves state-of-the-art performance, consistently outperforming nine well-established high-dimensional FS methods in terms of feature reduction and model performance metrics.

	\item The approach demonstrates strong generalization capabilities with low overfitting across tasks. Its multi-objective design balances exploration and exploitation effectively, ensuring scalability and reliability in high-dimensional contexts.
\end{enumerate}

The remainder of this article is organized as follows:
In Section \ref{sec:related}, we review related work in the field of feature selection, highlighting existing methods and their limitations in the context of high-dimensional data. Section \ref{sec:proposed} presents a detailed description of the proposed permutation-based subset evaluation feature selection method, outlining the underlying algorithm and its advantages. Section \ref{sec:experiments} covers the experiments conducted to evaluate the performance of our method, including the datasets used, experimental setup, and results. In Section \ref{sec:analysis}, we analyze the experimental results, discussing the implications and significance of our findings. Finally, Section \ref{sec:conclusions} concludes the article, summarizing the key contributions and suggesting directions for future research.

\section{Related works} 
\label{sec:related}
To provide a comprehensive understanding of the current landscape in FS for high-dimensional data, this section reviews the state-of-the-art methods developed in the previous and current year. The focus will be on the most recent advancements and approaches that have been proposed to address the challenges inherent in high-dimensional datasets, particularly in the context of improving model performance and computational efficiency.

Zhao \textit{et al.} \cite{ZHAO2024112042} introduce HMOFA, a hierarchical learning multi-objective firefly algorithm tailored for FS in high-dimensional datasets. The approach uses a clustering-based initialization method to reduce redundancy in the initial population while ensuring comprehensive exploration of the feature space. A hierarchy-guided learning model directs fireflies toward superior solutions and away from inferior ones, minimizing the risk of local optima and ineffective searches. Additionally, a duplicate solution modification mechanism enhances diversity by addressing redundancy in the population. Validated on 15 datasets, HMOFA demonstrates superior classification accuracy with fewer selected features compared to competitive methods.

Miao \textit{et al.} \cite{MIAO2024111979} propose RQWOA, a memory interaction quadratic interpolation whale optimization algorithm designed for FS in high-dimensional datasets. The method employs a novel quadratic interpolation mechanism with memory-based information interaction to enhance convergence speed and accuracy by improving datum point quality. Additionally, a reverse information correction mechanism leverages low-quality individuals to refine feature subsets, effectively reducing redundancy and removing irrelevant features. Validated on 15 high-dimensional datasets, RQWOA demonstrates an average accuracy improvement of 7.49\% and an 80.42\% reduction in selected features, showcasing its robust capability in feature identification and dimensionality reduction.

Jin \textit{et al.} \cite{JIN2024125084} propose Adaptive Pyramid PSO (APPSO) for high-dimensional FS, addressing challenges such as premature convergence and local optima in standard PSO. The approach introduces a weighted initialization strategy using feature correlation and a cubic chaotic map to enhance initial particle diversity. An adaptive constrained updating strategy, based on a pyramid structure, optimizes global search by assigning particles to different learning exemplars across pyramid layers. Additionally, a dynamic flip strategy (DFS) combines feature correlation and occurrence frequencies to strengthen local search capabilities. Experimental results on 18 datasets demonstrate APPSO's superior performance compared to eight wrapper-based FS methods.

Gong \textit{et al.} \cite{GONG2024111809} present a novel embedded FS approach for high-dimensional data based on the Takagi-Sugeno-Kang fuzzy system with a sparse rule base (TSK-SRB). The method introduces a refined softmin (Ref-softmin) function to approximate the minimum T-norm while avoiding arithmetic underflow, enabling accurate calculation of rule firing strengths. A broader initial rule base is generated using clustering to address the exponential growth of fuzzy rules in high dimensions. Additionally, automatic threshold segmentation is implemented to determine the optimal number of selected features and rules without requiring extra hyperparameters. Evaluated on 17 datasets, TSK-SRB demonstrates competitive performance and robustness in high-dimensional FS.

Feng \textit{et al.} \cite{FENG2024101618} propose MO-FSEMT, an evolutionary multitasking framework for multi-objective high-dimensional FS. The method constructs auxiliary tasks using filtering and clustering techniques to enhance knowledge diversity and transfer among tasks. It employs a multi-solver optimization scheme where each task is assigned independent populations and solvers with different search preferences, promoting solution diversity. A task-specific knowledge transfer mechanism further exploits elite solutions, enhancing feature masks and weights across tasks to guide optimization effectively. Evaluated on 27 high-dimensional datasets, MO-FSEMT outperforms state-of-the-art methods in both effectiveness and efficiency, demonstrating its robustness and capability in solving complex FS problems.

Yang \textit{et al.} \cite{YANG2024121185} propose HMOFS-CFDVD, a hybrid FS algorithm tailored for high-dimensional data, combining correlation-based filtering and a MOEA. The first stage employs a correlation filter for coarse-grained feature selection, eliminating redundancies and accelerating subsequent processing. The second stage refines the feature subset using a MOEA with dynamic sample variation distance, incorporating innovations such as co-evolutionary initialization, adaptive evolutionary strategies, and diversity mechanisms to enhance exploration and avoid local minima. Additionally, the algorithm optimizes three objectives: subset size, classification error, and a correlation metric. Tested on datasets with up to 21,655 features, HMOFS-CFDVD achieved competitive performance, producing compact subsets with high classification accuracy.

Bohrer and Dorn \cite{BOHRER2024124518} propose a hybrid FS approach using a multi-objective genetic algorithm to address challenges in high-dimensional data classification. The method combines outputs from classical FS techniques through innovative genetic operators, enabling exploration of optimized feature subsets beyond individual methods' capabilities. By leveraging the strengths of various selection methods, the approach narrows the search space and adapts to diverse datasets, enhancing classification accuracy while reducing dimensionality. Built on NSGA-II, the method ensures robust, domain-agnostic FS by exploring intermediate solutions and balancing performance metrics with dimensionality reduction. Experimental results validate its effectiveness across varied scenarios.

Ahadzadeh \textit{et al.} \cite{AHADZADEH2024101715} introduce UniBFS, a novel Uniform-solution-driven Binary FS algorithm tailored for addressing the challenges of high-dimensional data. UniBFS employs binary coding to conduct an exhaustive global search, effectively distinguishing relevant features while deactivating irrelevant ones. To enhance its performance, the authors propose the Redundant Features Elimination (RFE) algorithm, which performs a localized search within a reduced subspace derived from UniBFS solutions. RFE identifies and eliminates redundant features that do not contribute to classification accuracy. Additionally, a hybrid algorithm, UniBFS-ReliFish, integrates UniBFS with two filter-based FS methods--ReliefF and Fisher--to pinpoint critical features during the global search phase. Experimental validation on 30 datasets with dimensionalities ranging from 2000 to 54676 demonstrates that UniBFS and its variants outperform state-of-the-art techniques in both efficiency and effectiveness, offering robust solutions to high-dimensional FS challenges.

Peng \textit{et al.} \cite{PENG2024111734} introduce RLHHO, a Q-learning-guided mutational Harris Hawks Optimization algorithm, to address FS challenges in high-dimensional gene data. By integrating Q-learning with mutation strategies, RLHHO enhances global optimization and adapts to discrete combinatorial feature selection tasks. The authors also develop bRLHHO, a binary variant tailored for medical gene datasets, leveraging KNN for feature evaluation. Experiments on 12 high-dimensional datasets show RLHHO's superiority over traditional HHO, achieving higher classification accuracy and selecting fewer features with faster convergence. This work highlights RLHHO's efficacy in processing gene datasets exceeding 1000 dimensions.

Rehman \textit{et al.} \cite{REHMAN2024101701} propose a novel Reinforced Steering Evolutionary Markov Chain (RSEMC) framework for high-dimensional FS, integrating evolutionary algorithms, reinforcement learning, and Markov chains. The approach enhances communication among agents and guides them toward global optima using a reward-based reinforcement mechanism that adjusts transition probabilities dynamically. Improved agents are iteratively generated through the Markov process, limiting computational complexity by restricting the number of agents and focusing on the best-performing ones. The algorithm recursively reduces feature sets by retaining relevant features and eliminating non-contributing ones, ensuring efficient dimensionality reduction. Experimental results show that RSEMC outperforms state-of-the-art methods in accuracy and efficiency.

Li and Chai \cite{LI2024123296} introduce MPEA-FS, a decomposition-based multi-population evolutionary algorithm designed for high-dimensional FS. To address the challenge of balancing feature subset minimization with classification accuracy, the method employs a multi-population generation strategy (MGS) that ranks feature weights using Fisher scores and symmetric uncertainty, dividing the search space into smaller subspaces for focused exploration. An external population mechanism integrates optimal feature subsets from subpopulations, enabling knowledge sharing and enhancing diversity. Additionally, a feature reduction strategy based on a one-dimensional guide vector effectively removes irrelevant features while preserving accuracy. Experiments on 13 datasets show MPEA-FS achieves superior classification accuracy and competitive feature reduction compared to recent state-of-the-art methods.

Ma \textit{et al.} \cite{MA2024111948} introduce RWLLEA, a roulette wheel-based level learning evolutionary algorithm tailored for FS in high-dimensional datasets. The method utilizes a leveled population model where lower-level individuals learn from higher-level ones, enhancing diversity and exploiting feature interactions. Additionally, a roulette wheel-based update mechanism dynamically reduces the search space and balances exploration and exploitation throughout the evolutionary process. Evaluated on 15 diverse datasets, RWLLEA achieves superior classification accuracy with reduced feature subsets and lower computational costs, addressing the challenges of sparse and discrete optimization in high-dimensional data.

Abroshan and Moattar \cite{ABROSHAN2024111274} present DDFS-SLRBM, a novel discriminative FS method based on Signed Laplacian Restricted Boltzmann Machines (SLRBM). This approach is designed to enhance speed and generalization in the classification of high-dimensional data. Its main innovation lies in updating the weight matrix using a neighborhood matrix that considers similar and dissimilar classes, ensuring effective discrimination between them. Additionally, selected features are identified using the minimum reconstruction error criterion. Experiments conducted on datasets such as MNIST, GISETTE, and Protein demonstrate that DDFS-SLRBM improves classification accuracy, reduces processing time, and is scalable for different problem types and data sizes. The method automatically determines the optimal number of relevant features without predefined limitations, adapting intelligently to the data type.

Wang \textit{et al.} \cite{WANG2024120867} introduce a two-stage clonal selection algorithm (TSCSA-LFS) tailored for local FS in high-dimensional datasets. Unlike conventional global FS methods, TSCSA-LFS identifies feature subsets for distinct sample regions, accounting for local sample behaviors. The approach incorporates an enhanced discrete clonal selection algorithm with mutual information-based initialization, a mutation strategy pool leveraging differential evolution, and a local search technique, all aimed at improving search efficiency in discrete spaces. Additionally, it uses a two-part chromosome representation that automates parameter adjustments, minimizing manual tuning. Experimental results across 14 high-dimensional datasets demonstrate superior performance over established global and local FS methods, particularly in heterogeneous sample spaces.

Li \textit{et al.} \cite{LI2024120269} present an Enhanced NSGA-II (E-NSGA-II) method for FS in high-dimensional classification tasks, addressing challenges related to large, discrete decision spaces. The method integrates sparse initialization to reduce computational costs and maintain critical features, while a guided selection operator balances exploration and exploitation during the search process. Additionally, an innovative mutation strategy dynamically shrinks the search space, preventing convergence to local optima, and a greedy repair strategy refines feature subsets for improved classification performance. Validated on 15 high-dimensional datasets, E-NSGA-II outperforms eight competitive methods, delivering higher accuracy with smaller, less redundant feature subsets.

Miao \textit{et al.} \cite{MIAO2025112634} introduce the EMSWOA, a novel FS algorithm tailored for high-dimensional classification problems. The method incorporates a dynamic multi-swarm whale strategy guided by centroid distances to enhance subpopulation connectivity, improving both exploration and exploitation. A unique elite tuning mechanism corrects feature flipping errors caused by random thresholds, leveraging elite solutions to improve the recognition of relevant features. Additionally, dynamic population recombination and a local sparsity metric maintain candidate solution diversity and global search effectiveness. Validated on 19 datasets, EMSWOA achieved superior classification accuracy and significantly reduced feature subsets compared to six benchmark algorithms.

Li \textit{et al.} \cite{LI2025112574} propose the AMFEA, an adaptive multifactor evolutionary algorithm for FS in high-dimensional classification problems. The method employs a multi-task optimization framework, generating distinct tasks via filter-based methods and leveraging knowledge transfer to enhance classification performance. An adaptive parameter matrix dynamically adjusts knowledge transfer probabilities to balance positive and negative transfers, improving efficiency. Additionally, a local search strategy is incorporated to avoid local optima and identify optimal feature subsets. Evaluated on 18 high-dimensional datasets, AMFEA outperforms traditional FS methods and evolutionary algorithms in terms of classification accuracy.

\subsection*{Conclusions of related works}

The review of state-of-the-art methods highlights significant advances in feature selection for high-dimensional data, including evolutionary approaches, hybrid techniques, and decomposition-based frameworks. These methods employ innovative strategies such as multi-population mechanisms, task-specific knowledge transfer, adaptive updating strategies, and novel scoring metrics to address the challenges posed by high-dimensional datasets. While many of these approaches are efficient and scalable, they often involve complex optimization or rely on heuristics that, in some cases, may introduce computational overhead or require specific adjustments for different data scenarios.

In this context, the proposed PSEFS-MOEA method introduces a paradigm shift by leveraging a novel permutation-based evaluation concept for feature subsets. Unlike other approaches, PSEFS-MOEA eliminates the need for training machine learning models iteratively during the feature selection process. Instead, it relies on a baseline model trained offline, significantly reducing computational costs while ensuring scalability. This unique characteristic makes it particularly well-suited for high-dimensional data, providing a robust and efficient alternative to existing methods. Additionally, the method is easy to implement and offers versatility, being applicable not only to balanced classification tasks but also to imbalanced classification and regression problems, without requiring modifications to its implementation for different tasks. Furthermore, its multi-objective nature ensures a balanced trade-off between exploration and exploitation during the search process, enabling the discovery of optimal feature subsets across diverse predictive tasks.

\section{Permutation feature importance}
\label{sec:PFI}
PFI is a model-agnostic method widely used to estimate the contribution of individual features to the performance of a predictive model. PFI was initially introduced by Breiman \cite{Breiman2001} for use with \textit{random forests}. Building on this concept, Fisher \textit{et al.} \cite{DBLP:journals/jmlr/FisherRD19} extended the idea to create a model-agnostic approach, which they termed \textit{model reliance}. PFI evaluates how much the prediction error of the model increases when the values of a particular feature are randomly shuffled while keeping the other features intact. This shuffling disrupts the relationship between the selected feature and the target variable, allowing us to measure the decrease in the model's performance caused by the loss of information from that feature.

The process of computing PFI typically involves the following steps:

\begin{enumerate}
	\item \textit{Baseline performance measurement}: Calculate the initial prediction error, on an evaluation dataset $E$, of the trained model. 
	
	\item \textit{Feature permutation}: For each feature, shuffle its values across all samples in the evaluation dataset $E$ to break its relationship with the target variable.

	\item \textit{Performance evaluation}: Recalculate the model's prediction error using the evaluation dataset $E$ with the shuffled feature.
	
	\item \textit{Importance calculation}: Compute the difference between the baseline error and the error with the permuted feature. A larger increase in error indicates a more important feature.
\end{enumerate}

PFI offers several advantages. It is model-agnostic, meaning it can be applied to any predictive model regardless of the underlying algorithm, making it a highly versatile tool for feature evaluation. Its interpretability is another key strength, as the importance scores are directly linked to changes in model performance, providing an intuitive understanding of feature relevance. Furthermore, because the model's prediction error reflects the combined effects of all features, permutation feature importance inherently captures multivariate interactions, allowing it to account for relationships between features that influence the model's performance.

However, PFI also has some limitations. It assumes independence between the permuted feature and the others, which can lead to misleading importance scores for highly correlated features. Although the method avoids retraining the model, evaluating importance for multiple features can still be computationally expensive, especially for large datasets. The results may also vary due to random permutations and data distribution, requiring repeated evaluations to obtain reliable estimates. Additionally, traditional PFI evaluates features individually, which may overlook complex interactions among subsets of features.

There is an ongoing debate regarding whether feature importance should be computed using the training or validation data as the evaluation dataset $E$. Computing on the training data reflects the features' importance as perceived by the model during training, which can be useful for understanding how the model fits the data. However, it may overestimate feature importance if the model has overfitted to the training data.

On the other hand, computing feature importance on the validation data provides a measure of generalization, offering insights into how features contribute to predictions on unseen data. This is often considered more desirable for real-world applications. However, as highlighted in \cite{molnar2022}, the choice between training and validation data is not straightforward, as both perspectives offer valuable insights.

To address this uncertainty, we explore both approaches in our experiments. By evaluating feature importance on both the training and validation datasets, we aim to provide a comprehensive understanding of the features' relevance during training and their generalization to unseen data, highlighting the strengths and limitations of this approach across different perspectives. This also sets the stage for the proposed method, which extends permutation feature importance to evaluate subsets of features more effectively.

\section{A multi-objective evolutionary algorithm for permu\-tation-based subset evaluation feature selection}
\label{sec:proposed}
In this section, we introduce the proposed method, PSEFS-MOEA, which employs a MOEA to search for optimal feature subsets. The method aims to maximize the performance degradation of a pre-trained model, where the feature values of the selected subset are permuted, while simultaneously minimizing the cardinality of the subset. In Section \ref{sec:problem}, we formulate the proposed subset evaluation FS problem as a multi-objective boolean combinatorial optimization problem, presenting two versions of the problem set up for consideration. Section \ref{sec:MOEA} provides a description of the MOEA and its characteristics.

\subsection{Problem formulation}
\label{sec:problem}
We consider two versions of the problem formulation, computing the feature importance on the validation  data and the training data, respectively. Let $D=\{\bm{d}_1,\ldots,\bm{d}_s\}$ be a dataset with $s$ instances. Each instance or sample $\bm{d}_t=\{d_t^1,\ldots,d_t^w,o_t\}$, $t=1,\ldots,s$, has $w$ input attributes of any type,
and one output attribute $o_t\in\mathds{R}$, for regression problems, or $o_t\in\{S_1,\ldots,S_q\}$, $q>1$, for classification problems, where $S_k$, $k=1,\ldots,q$, are categorical output classes.
Dataset $D$ is divided into three partitions $R$, $V$ and $T$ for training, validation  and test with $60\%$, $20\%$  and $20\%$ of the data respectively. We consider the two following \textit{multi-objective combinatorial optimization problems}:

\begin{equation}
\label{eq:v1}
\begin{array}{ll}
Maximize & f_1(\textbf{x})=\mathcal{F}(\textbf{x},M_R^\Phi,V) \\
Minimize & f_2(\textbf{x})=\mathcal{C}(\textbf{x}) \\
\end{array}
\end{equation}

\begin{equation}
\label{eq:v2}
\begin{array}{ll}
Maximize & f_1(\textbf{x})=\mathcal{F}(\textbf{x},M_Q^\Phi,Q) \\
Minimize & f_2(\textbf{x})=\mathcal{C}(\textbf{x}) \\
\end{array}
\end{equation}

\noindent 
where $\textbf{x} = (x_1,\ldots,x_w)\in\{0,1\}^w$ is the decision variable set. Each decision variable $x_i$, $i=1,\ldots,w$, indicates whether the input attribute $i$ is selected ($x_i=1$) or not
($x_i=0$) in the feature selection process. Function $\mathcal{F}$ is the degradation in performance of a previously trained model $ M $ with all $w$ attributes when feature subset $ \textbf{x} $ values are permuted and evaluated on an evaluation set $ E $ using some performance metric, as described in Algorithm \ref{alg:fitness}. In problem (\ref{eq:v1}), the model $M$ is trained with the training dataset $R$ and evaluated using the validation dataset $V$, i.e., $M=M_R^\Phi$ and $E=V$, where $\Phi$ is some learning algorithm\footnote{We used \textit{Random Forest} (RF) in the experiments.}. In problem (\ref{eq:v2}), the model $M$ is trained and evaluated with the same dataset $Q=R\cup V$, i.e., $M=M_Q^\Phi$ and $E=Q$. The function $\mathcal{C}$ is the cardinality of the feature subset represented by $ \textbf{x} $, that is, the number of variables $ x_i $ such that $ x_i = 1 $, $i=1,\ldots,w$.

\begin{algorithm}[!htpb]
	\caption{Function $\mathcal{F}(\textbf{x},M,E)$}
	\label{alg:fitness}
	\begin{algorithmic}[1]
		\REQUIRE $\textbf{x}=\{x_1,\ldots,x_w\}$  \COMMENT{Decision variable set}
		\REQUIRE {$M$} \COMMENT{Prediction model previously trained with a dataset with all attributes}
		\REQUIRE $E\subset D$ \COMMENT{{Evaluation dataset}}
		\STATE $A\leftarrow P(M,E)$ \COMMENT{Baseline performance measurement}
		\FOR {$i=1$ \TO $w$}
		\IF {$x_i=1$}
		\STATE shuffle($E$, $i$) \COMMENT{Feature permutation}
		\ENDIF
		\ENDFOR
		\STATE $B\leftarrow P(M,E)$ \COMMENT{Performance evaluation}
		\RETURN $\big|A-B\big|$ \COMMENT{Merit of $ \textbf{x} $}
	\end{algorithmic}
\end{algorithm}

To calculate the function $\mathcal{F}$ with the formal parameters $ \textbf{x} $, $ M $ and $ E $, Algorithm \ref{alg:fitness} performs the following steps: first,  the performance of the model $ M $, evaluated on the dataset $ E $, is calculated using a performance metric\footnote{In this paper we use \textit{accuracy} (ACC) for classification problems and \textit{root mean square error} (RMSE) for regression problems} $P$  (line 1); then the values of the attributes in $ E' $ that have been selected in $ \textbf{x} $, i.e. $x_i=1$, $i=1,\ldots,w$, are shuffled (lines 2 to 6). Next, the performance of the model $ M $ is recalculated using the evaluation data set $E$ with the shuffled features and the performance metric $ P $ (line 7). Finally, the absolute value of the difference between the baseline performance and the performance with the permuted features is returned (line 8).

\subsection{Multi-objetive evolutionary algorithm}
\label{sec:MOEA}	
For the proposed PSEFS-MOEA method, we employed the \textit{Non-dominated Sorting Genetic Algorithm II} (NSGA-II) \cite{Deb02} with binary representation to solve the optimization problems (\ref{eq:v1}) and (\ref{eq:v2}) defined in Section \ref{sec:problem}. NSGA-II is a widely used multi-objective evolutionary algorithm designed to optimize problems with conflicting objectives. The algorithm operates by maintaining a diverse population and employing a fast non-dominated sorting approach to classify solutions into Pareto fronts. It also uses a crowding distance mechanism to ensure diversity among solutions. The popularity of NSGA-II stems from its efficiency and robustness in handling multi-objective optimization tasks. Although NSGA-II was chosen for its well-established performance and accessibility, other multi-objective optimization algorithms could also have been used.

The fitness functions correspond to the objective functions $ f_1 $ and $ f_2 $ 
outlined in the optimization problems (\ref{eq:v1}) and (\ref{eq:v2}). Since the function $ f_1 $ is a maximization objective and  $ f_2 $ is a minimization objective,  $ f_1 $
has been multiplied by $-1$ so that both  $f_1$ and 
$ f_2 $ are treated as minimization objectives in the MOEA implementation. To evolve the population, we utilized \textit{half uniform crossover} \cite{ESHELMAN1991}  and \textit{bit flip mutation} \cite{davis1991} operators. After running the algorithm, the solution with the best $ f_1(\textbf{x}) $
value was selected from the Pareto front identified by NSGA-II. To implement our method, we leveraged the \textit{Platypus} platform \cite{platypus}, a versatile library for evolutionary computation and multi-objective optimization, which facilitated the integration of NSGA-II with our problem-specific requirements.

\section{Experiments and results} 
\label{sec:experiments}
In this section, we describe the experiments conducted to evaluate the performance of the proposed PSEFS-MOEA method. The general goal is to assess its ability to identify effective feature subsets for both classification and regression tasks, comparing its performance against several established FS methods across a variety of scenarios and datasets.

We first provide details of the datasets used in the experiments, which include a variety of high-dimensional datasets from classification and regression domains (Section \ref{sec:datasets}). Next, we describe the comparison methods used as benchmarks to evaluate PSEFS-MOEA (Section \ref{sec:methods}).  Finally, Section \ref{sec:results} presents the results of the experiments, while the analysis and discussion of these results are deferred to Section \ref{sec:analysis}.

\subsection{Datasets}
\label{sec:datasets}
The experiments were conducted on a total of 24 datasets for classification and regression tasks. The number of features in these datasets ranges from a minimum of 617 to a maximum of 22,277, ensuring a diverse set of high-dimensional scenarios for evaluation.

The 14 classification datasets were obtained from the public \textit{OpenML} platform\footnote{\url{https://www.openml.org/}}. These datasets vary in the number of instances and include both binary and multi-class classification problems. The class distributions also vary, covering both balanced and imbalanced scenarios. A summary of the key characteristics of the classification datasets is provided in Table \ref{tab:classification}.

\begin{table}[!h]
	\centering
	\resizebox{0.48\textwidth}{!}{
	\begin{tabular}{lrrrc}
		\hline
		\textbf{Name} & \textbf{Instances} & \textbf{Attributes} & \textbf{Classes}  & \textbf{Balanced} \\ \hline    
		isolet & 600 & 617 & 2 & Yes \\                    
		cnae-9 & 240 & 856 & 2  & Yes \\
		micro-mass & 571 & 1,300 & 20 & No \\ 
		Colon & 62 & 2,000 & 2 & No \\
		lymphoma\_11classes & 96 & 4,026 & 11 & No \\
		eating & 945 & 6,373 & 7 & No \\		
		tr45.wc & 690 & 8,261 & 10 & No \\		
		amazon-commerce-reviews	& 1,500 & 10,000 & 50 & Yes \\
		AP\_Colon\_Kidney & 546 & 10,936 & 2 & No \\
		11\_tumors & 174 & 12,533 & 11 & No \\		
		Lung & 203 & 12,600 & 5 & No \\			
		GCM & 190 & 16,063 & 14 & No \\
		rsctc2010\_3 & 95 & 22,277 & 5 & No \\	
		BurkittLymphoma & 220 & 22,283 & 3 & No \\								
		\hline
	\end{tabular}
	}
	\caption{Datasets for classification, sorted from lowest to highest number of attributes.}
	\label{tab:classification}
\end{table}

For regression tasks, we used 10 datasets, with five obtained from the \textit{OpenML} platform and the other five generated synthetically. The synthetic datasets were designed with varying numbers of instances, attributes, informative attributes, and levels of noise to introduce different levels of difficulty. Table \ref{tab:regression} provides an overview of the characteristics of these regression datasets.

\begin{table}[!h]
	\centering
	\resizebox{0.48\textwidth}{!}{
		\begin{tabular}{lrrrr}
			\hline
			\textbf{Name} & \textbf{Instances} & \textbf{Attributes} & \textbf{Inf. attr.}  & \textbf{Noise} \\ \hline    
			synthetic\_dataset1 & 1,000 & 1,000 & 500 & 0.1 \\	
			QSAR-TID-11054  & 344 & 1,024 & -- & -- \\
			QSAR-TID-11110  & 969 & 1,024 & -- & -- \\
			QSAR-TID-11524  & 605 & 1,024 & -- & -- \\
			mtp2 & 274 & 1,142 & -- & -- \\	
			synthetic\_dataset2 & 900 & 2,000 & 1,000 & 0.2 \\	
			synthetic\_dataset3 & 800 & 3,000 & 1,500 & 0.3 \\
			liverTox\_ALT\_target & 64 & 3,116 & -- & --\\
			synthetic\_dataset4 & 700 & 4,000 & 2,000 & 0.4 \\	
			synthetic\_dataset5 & 600 & 5,000 & 2,500 & 0.5 \\										
			\hline
		\end{tabular}
	}
	\caption{Datasets for regression, sorted from lowest to highest number of attributes.}
	\label{tab:regression}
\end{table}

\subsection{Comparison methods}
\label{sec:methods}
The proposed feature selection method for high-dimensional data has been compared against nine established methods, encompassing both subset evaluation and attribute evaluation approaches. All comparison methods are filter-based and include both multivariate and univariate techniques. Wrapper methods were excluded from this study due to their high computational cost and impracticality when dealing with high-dimensional datasets.

The proposed feature selection method, PSEFS-MOEA, is evaluated in two versions, referred to in this paper as PSEFS-MOEA-V1 and PSEFS-MOEA-V2, which correspond to the problem formulations (\ref{eq:v1}) and (\ref{eq:v2}), respectively. For classification tasks, the following comparison methods were considered:

\begin{enumerate}
\item CSE-BF: This method uses the \textit{ConsistencySubsetEval} algorithm in \textit{Weka}\footnote{\url{https://ml.cms.waikato.ac.nz/weka}} \cite{Weka}, which evaluates feature subsets based on their consistency with the target variable \cite{liu1996probabilistic}. The subset search strategy is performed using a \textit{Best First} (BF) \cite{10.5555/2974989} approach.

\item CFSSE-BF: The \textit{CfsSubsetEval} algorithm in \textit{Weka} evaluates feature subsets based on their predictive ability and redundancy \cite{hall1999correlation,hall2000correlation}. It selects subsets that are highly correlated with the target variable but have low inter-correlation. The search is also performed using a BF strategy.

\item PFI-V1: This is the PFI method\footnote{\url{https://scikit-learn.org/1.5/modules/permutation\_importance.html}} implemented in \textit{scikit-learn} \cite{scikit-learn,sklearnapi} and described in Section \ref{sec:PFI}, which evaluates the importance of individual features by measuring the increase in prediction error when the feature values are permuted. In this version, the evaluation is performed using a validation dataset.

\item PFI-V2: Similar to PFI-V1, this version of PFI evaluates feature importance using the training dataset instead of the validation dataset.

\item SAE: The \textit{SignificanceAttributeEval} method in \textit{Weka} assesses the significance of individual attributes based on statistical significance tests \cite{Ahm2005}.

\item IGAE: The \textit{InfoGainAttributeEval} method in \textit{Weka} evaluates individual features by calculating the information gain with respect to the target variable, identifying the attributes that provide the most reduction in entropy \cite{Karegowda2010}.

\item RFAE: The \textit{ReliefFAttributeEval} method in \textit{Weka} is a multivariate technique that assesses feature importance by measuring the ability of each feature to distinguish between instances that are near in the feature space but belong to different classes \cite{kononenko1994estimating}.

\item CSAE: The \textit{ChiSquaredAttributeEval} method in \textit{Weka} uses the Chi-squared test to evaluate the dependency between individual features and the target variable, ranking features based on their Chi-squared scores \cite{galavotti2000experiments}.

\item CAE: The \textit{CorrelationAttributeEval} method in \textit{Weka} evaluates features based on their Pearson correlation coefficient with the target variable, ranking features with higher absolute correlations as more important \cite{freedman2007statistics}.
\end{enumerate}

For regression tasks, the PSEFS-MOEA-V1, PSEFS-MOEA-V2, CFS-BF, PFI-V1, PFI-V2, RFAE, and CAE methods were considered in the experiments 

\subsection{Description of the experiments}
\label{sec:descexp}
 Specifically, the experiments are designed to address the following objectives:

\begin{itemize}
	\item \textit{Comparison of PSEFS-MOEA versions}: To compare the two proposed variants, PSEFS-MOEA-V1 and PSEFS-MOEA-V2, in terms of their performance on the optimization problems (\ref{eq:v1}) and (\ref{eq:v2}) described earlier.

\item \textit{Evaluation against conventional PFI}: To determine whether the proposed permutation-based method for subset evaluation outperforms the conventional PFI method, which evaluates features individually, particularly in capturing feature interactions and improving predictive performance.

\item \textit{Effectiveness in reducing feature sets}: To assess the ability of the proposed method to reduce the number of selected features while maintaining or enhancing the predictive accuracy of the models.

\item \textit{Comparison with established methods}: To compare the proposed method with other well-established feature selection techniques for high-dimensional data. This includes evaluating predictive performance, overfitting analysis and computational efficiency in terms of execution time.

\item \textit{Robustness across tasks}: To validate the robustness of the proposed method in both classification and regression tasks, including imbalanced classification problems, by employing appropriate performance metrics for each type of task.
\end{itemize}

With these objectives and premises in mind, the  experiments were conducted as follows:

\begin{itemize}
	\item The proposed PSEFS-MOEA method is probabilistic, requiring multiple runs with different random seeds to ensure reliable and statistically sound results. Both versions PSEFS-MOEA-V1 and PSEFS-MOEA-V2 of the proposed method were executed 10 times with different random seeds across the 24 datasets considered in this study. RF was used to construct the model for evaluating feature importance. For V1, the model was trained on the dataset $R$ (60\% of the data), and features were evaluated on the validation set $V$ (20\% of the data), following the problem formulation in Eq. (\ref{eq:v1}). For V2, the model was trained and evaluated on the dataset $Q=R\cup V$ (80\% of the data), following the formulation in Eq. (\ref{eq:v2}). The parameter settings used in the executions, as well as those for the comparison methods, are summarized in Table \ref{tab:param}.

\begin{table}[!h]
	\centering
	\resizebox{0.48\textwidth}{!}{
		\begin{tabular}{ll}
			\hline
			\textbf{Method} & \textbf{Parameters} \\ \hline    
			PSEFS-MOEA-V1 & $M=M_R^\Phi$, $\Phi=$ RF, $E=V$, \\
			& population size = 50, generations = 2,000, \\
			& crossover probability = 1.0, mutation probability = 0.02 \\
			
			PSEFS-MOEA-V2 & $M=M_Q^\Phi$, $\Phi=$ RF, $E=Q$, \\
			& population size = 50, generations = 2,000, \\
			& crossover probability = 1.0, mutation probability = 0.02 \\
			
			CSE-BF & weka.filters.supervised.attribute.AttributeSelection \\
			& -E ``weka.attributeSelection.ConsistencySubsetEval" \\
			& -S ``weka.attributeSelection.BestFirst -D 1 -N 5"\\
			
			CFSSE-BF & weka.filters.supervised.attribute.AttributeSelection \\
			& -E ``weka.attributeSelection.CfsSubsetEval -P 1 -E 1" \\
			& -S ``weka.attributeSelection.BestFirst -D 1 -N 5"\\	
			
			PFI-V1 & model $=M_R^\Phi$, $\Phi=$ RF, (X, y) $= V $, n\_repeats$=5$ \\
			
			PFI-V2 & model $=M_Q^\Phi$, $\Phi=$ RF, (X, y) $= Q $, n\_repeats$=5$ \\
			
			SAE & weka.filters.supervised.attribute.AttributeSelection \\
			& -E  ``weka.attributeSelection.SignificanceAttributeEval"\\
			& -S ``weka.attributeSelection.Ranker -T -1.8E308 -N -1"\\
			
			IGAE & weka.filters.supervised.attribute.AttributeSelection \\
			& -E ``weka.attributeSelection.InfoGainAttributeEval" \\
			& -S ``weka.attributeSelection.Ranker -T -1.8E308 -N -1"\\
			
			RFAE & weka.filters.supervised.attribute.AttributeSelection \\
			& -E ``weka.attributeSelection.ReliefFAttributeEval \\
			&  -M -1 -D 1 -K 10" \\
			& -S ``weka.attributeSelection.Ranker -T -1.8E308 -N -1"\\
			
			CSAE & weka.filters.supervised.attribute.AttributeSelection \\
			& -E ``weka.attributeSelection.ChiSquaredAttributeEval" \\
			& -S ``weka.attributeSelection.Ranker -T -1.8E308 -N -1"\\
			
			CAE & weka.filters.supervised.attribute.AttributeSelection \\
			& -E ``weka.attributeSelection.CorrelationAttributeEval" \\
			& -S ``weka.attributeSelection.Ranker -T -1.8E308 -N -1"\\
			\hline
		\end{tabular}
	}
	\caption{Parameters of the feature selection methods used in this paper.}
	\label{tab:param}
\end{table}

\item Since PFI also requires a pre-trained model for evaluating feature importance, it was executed in two versions V1 and V2. In V1, the model was trained on the dataset $R$ and features were evaluated on a separate validation set $V$, while in V2 the model was trained on the dataset $Q=R\cup V$ and features ware evaluated on the same dataset $Q$. The model was built using the RF learning algorithm in both versions.

\item In contrast, the other FS methods do not use pre-trained models, evaluating features directly on the dataset $Q=R\cup V$.  For these FS methods, it is necessary to first generate a ranking of features and then build models using the top-ranked features for a predefined number of attributes. In our experiments, models were constructed with the top 10, 100, $N1$ and $N2$
features. $N1$ and $N2$ are the number of features selected by PSEFS-MOEA-V1 and PSEFS-MOEA-V2, respectively, to ensure a fair comparison.

\item To evaluate the quality of the feature subsets selected by each method, classification or regression models (depending on the task) were built using the RF algorithm on the training dataset 
$Q=R\cup V$ (80\% of the data). The resulting models were then evaluated on both the training dataset 
$Q$ and an unseen test dataset $T$ (20\% of the data). The test dataset $T$ was not exposed to any FS algorithm at any point. Evaluating models on both training and test datasets enables an analysis of overfitting and the generalization capability of the models produced with each FS method.

\item The performance of classification models was assessed using ACC and \textit{balanced accuracy} (BA), while for regression models, RMSE and \textit{coefficient of determination} ($R^2$) were employed. ACC provides a straightforward measure of classification performance but can be misleading in imbalanced datasets. BA adjusts for imbalanced datasets by averaging \textit{recall} across classes.
RMSE captures the magnitude of prediction errors in regression. $R^2$ measures the proportion of variance in the target variable explained by the model.
These metrics provide a comprehensive evaluation of the predictive performance of the models across different tasks.

\item For subset evaluation methods (PSEFS-MOEA-V1, PSEFS-MOEA-V2, CSE-BF, and CFS-BF), the number of selected features was recorded. Additionally, the runtime of all feature selection methods was measured to assess their computational efficiency.

\end{itemize}

\subsection{Results}
\label{sec:results}

The results obtained from the experiments are summarized in the following tables:

\begin{itemize}
	\item Table \ref{tab:ACCTrain} reports the accuracy of the models evaluated on the training set 
$Q=R\cup V$ for classification problems.

\item Table \ref{tab:ACCTest} reports the accuracy of the models evaluated on the test set $T$ for classification problems.

\item Table \ref{tab:BATrain} presents the balanced accuracy of the models evaluated on the training set 
$Q=R\cup V$ for classification problems.

\item Table \ref{tab:BATest} presents the balanced accuracy of the models evaluated on the test set 
$T$ for classification problems.

\item  Table \ref{tab:RMSETrain} shows the RMSE of the models evaluated on the training set 
$Q=R\cup V$ for regression problems.

\item Table \ref{tab:RMSETest} shows the RMSE of the models evaluated on the test set 
$T$ for regression problems.

\item Table \ref{tab:R2Train} reports the 
$R^2$ of the models evaluated on the training set 
$Q=R\cup V$ for regression problems.

\item Table \ref{tab:R2Test} reports the 
$R^2$ of the models evaluated on the test set 
$T$ for regression problems.

\item Tables \ref{tab:RunTimesClassification} and \ref{tab:RunTimesRegression} display the runtime of the FS methods for classification and regression problems, respectively. For the PSEFS-MOEA-V1 and PSEFS-MOEA-V1 methods, the average time of the 10 runs for each dataset is reported. 
\end{itemize}

\begin{table*}[!h]
	\begin{center}
		\resizebox{\textwidth}{!}{
\begin{tabular}{llrrrrrrrrrrrrrr}\hline
				\textbf{Method} & \textbf{Metric} 
				& \textbf{amazon-commerce-reviews}
				& \textbf{AP-Colon-Kidney}
				& \textbf{BurkittLymphoma}
				& \textbf{cnae-9}
				& \textbf{Colon}
				& \textbf{eating}
				& \textbf{isolet}
				& \textbf{Lung}
				& \textbf{lymphoma-11classes}
				& \textbf{micro-mass}
				& \textbf{11-tumors}
				& \textbf{GCM}
				& \textbf{tr45.wc}
				& \textbf{rsctc2010-3}
				\\ \hline
				\multirow{2}{*}{All features} & No. attr. ($w$) 
				& $10000$ & $10936$ & $22283$ & $856$ & $2000$ & $6373$ & $617$ & $12600$ & $4026$ & $1300$ & $12533$ & $16063$ & $8261$ & $22277$\\
				& ACC & $\bf{1.0000} $ & $\bf{1.0000} $ & $\bf{1.0000} $ & $\bf{1.0000} $ & $\bf{1.0000} $ & $\bf{1.0000} $ & $\bf{1.0000} $ & $\bf{1.0000} $ & $\bf{1.0000} $ & $\bf{1.0000} $ & $\bf{1.0000} $ & $\bf{1.0000} $ & $\bf{1.0000} $ & $\bf{1.0000} $\\
				\hline
				\multirow{2}{*}{PSEFS-MOEA-V1} & No. attr. ($N1$) 
				& 1018
				& 942
				& 4929
				& 14
				& 20
				& 469
				& 11
				& 1446
				& 69
				& 68
				& 1790
				& 2665
				& 485
				& 5006
				\\
				& ACC & $\bf{1.0000} $ & $\bf{1.0000} $ & $\bf{1.0000} $ & $0.9635 $ & $\bf{1.0000} $ & $\bf{1.0000} $ & $\bf{1.0000} $ & $\bf{1.0000} $ & $\bf{1.0000} $ & $\bf{1.0000} $ & $\bf{1.0000} $ & $\bf{1.0000} $ & $\bf{1.0000} $ & $\bf{1.0000} $\\
				\hline
				\multirow{2}{*}{PSEFS-MOEA-V2} & No. attr. ($N2$) 
				& 1653
				& 994
				& 6560
				& 32
				& 57
				& 1168
				& 14
				& 1626
				& 281
				& 159
				& 1801
				& 4890
				& 694
				& 5131
				\\
				& ACC & $\bf{1.0000} $ & $\bf{1.0000} $ & $\bf{1.0000} $ & $0.9896 $ & $\bf{1.0000} $ & $\bf{1.0000} $ & $\bf{1.0000} $ & $\bf{1.0000} $ & $\bf{1.0000} $ & $\bf{1.0000} $ & $\bf{1.0000} $ & $\bf{1.0000} $ & $\bf{1.0000} $ & $\bf{1.0000} $\\
				\hline
				\multirow{2}{*}{CSE-BF} & No. attr. ($N3$) 
				& 35
				& 5
				& 5
				& 9
				& 3
				& 12
				& 4
				& 5
				& 5
				& 21
				& 7
				& 14
				& 16
				& 4
				\\
\hline
			\end{tabular}}
			\caption{Accuracy with Random Forest on the train dataset for the classification problems.}
			\label{tab:ACCTrain}
		\end{center}
	\end{table*}

\begin{table*}[!h]
	\begin{center}
		\resizebox{\textwidth}{!}{\begin{tabular}{llrrrrrrrrrrrrrr}\hline
				\textbf{Method} & \textbf{Metric} 
				& \textbf{amazon-commerce-reviews}
				& \textbf{AP-Colon-Kidney}
				& \textbf{BurkittLymphoma}
				& \textbf{cnae-9}
				& \textbf{Colon}
				& \textbf{eating}
				& \textbf{isolet}
				& \textbf{Lung}
				& \textbf{lymphoma-11classes}
				& \textbf{micro-mass}
				& \textbf{11-tumors}
				& \textbf{GCM}
				& \textbf{tr45.wc}
				& \textbf{rsctc2010-3}
				\\ \hline
				\multirow{2}{*}{All features} & No. attr. ($w$) 
				& $10000$ & $10936$ & $22283$ & $856$ & $2000$ & $6373$ & $617$ & $12600$ & $4026$ & $1300$ & $12533$ & $16063$ & $8261$ & $22277$\\
				& ACC & $0.6933 $ & $0.9725 $ & $0.8182 $ & $\bf{1.0000} $ & $\bf{0.8333} $ & $0.5238 $ & $\bf{0.9917} $ & $0.9750 $ & $0.7368 $ & $0.9123 $ & $0.8824 $ & $0.7368 $ & $0.9203 $ & $0.8947 $\\
				\hline
				\multirow{2}{*}{PSEFS-MOEA-V1} & No. attr. ($N1$) 
				& 1018
				& 942
				& 4929
				& 14
				& 20
				& 469
				& 11
				& 1446
				& 69
				& 68
				& 1790
				& 2665
				& 485
				& 5006
				\\
				& ACC & $0.7533 $ & $0.9725 $ & $0.8864 $ & $0.9583 $ & $\bf{0.8333} $ & $0.5979 $ & $0.9833 $ & $\bf{1.0000} $ & $\bf{0.8421} $ & $0.8947 $ & $0.8824 $ & $0.7368 $ & $0.9420 $ & $\bf{0.9474} $\\
				\hline
				\multirow{2}{*}{PSEFS-MOEA-V2} & No. attr. ($N2$) 
				& 1653
				& 994
				& 6560
				& 32
				& 57
				& 1168
				& 14
				& 1626
				& 281
				& 159
				& 1801
				& 4890
				& 694
				& 5131
				\\
				& ACC & $\bf{0.7767} $ & $0.9817 $ & $0.8636 $ & $0.9792 $ & $\bf{0.8333} $ & $\bf{0.6032} $ & $0.9833 $ & $\bf{1.0000} $ & $\bf{0.8421} $ & $\bf{0.9211} $ & $0.9118 $ & $\bf{0.7895} $ & $\bf{0.9493} $ & $0.8947 $\\
				\hline
				\multirow{2}{*}{CSE-BF} & No. attr. ($N3$) 
				& 35
				& 5
				& 5
				& 9
				& 3
				& 12
				& 4
				& 5
				& 5
				& 21
				& 7
				& 14
				& 16
				& 4
				\\
\hline
			\end{tabular}}
			\caption{Accuracy with Random Forest on the test dataset for the classification problems.}
			\label{tab:ACCTest}
		\end{center}
	\end{table*}
		
\begin{table*}[!h]
	\begin{center}
		\resizebox{\textwidth}{!}{\begin{tabular}{llrrrrrrrrrrrrrr}\hline
				\textbf{Method} & \textbf{Metric} 
				& \textbf{amazon-commerce-reviews}
				& \textbf{AP-Colon-Kidney}
				& \textbf{BurkittLymphoma}
				& \textbf{cnae-9}
				& \textbf{Colon}
				& \textbf{eating}
				& \textbf{isolet}
				& \textbf{Lung}
				& \textbf{lymphoma-11classes}
				& \textbf{micro-mass}
				& \textbf{11-tumors}
				& \textbf{GCM}
				& \textbf{tr45.wc}
				& \textbf{rsctc2010-3}
				\\ \hline
				\multirow{2}{*}{All features} & No. attr. ($w$) 
				& $10000$ & $10936$ & $22283$ & $856$ & $2000$ & $6373$ & $617$ & $12600$ & $4026$ & $1300$ & $12533$ & $16063$ & $8261$ & $22277$\\
				& BA & $\bf{1.0000} $ & $\bf{1.0000} $ & $\bf{1.0000} $ & $\bf{1.0000} $ & $\bf{1.0000} $ & $\bf{1.0000} $ & $\bf{1.0000} $ & $\bf{1.0000} $ & $\bf{1.0000} $ & $\bf{1.0000} $ & $\bf{1.0000} $ & $\bf{1.0000} $ & $\bf{1.0000} $ & $\bf{1.0000} $\\
				\hline
				\multirow{2}{*}{PSEFS-MOEA-V1} & No. attr. ($N1$) 
				& 1018
				& 942
				& 4929
				& 14
				& 20
				& 469
				& 11
				& 1446
				& 69
				& 68
				& 1790
				& 2665
				& 485
				& 5006
				\\
				& BA & $\bf{1.0000} $ & $\bf{1.0000} $ & $\bf{1.0000} $ & $0.9635 $ & $\bf{1.0000} $ & $\bf{1.0000} $ & $\bf{1.0000} $ & $\bf{1.0000} $ & $\bf{1.0000} $ & $\bf{1.0000} $ & $\bf{1.0000} $ & $\bf{1.0000} $ & $\bf{1.0000} $ & $\bf{1.0000} $\\
				\hline
				\multirow{2}{*}{PSEFS-MOEA-V2} & No. attr. ($N2$) 
				& 1653
				& 994
				& 6560
				& 32
				& 57
				& 1168
				& 14
				& 1626
				& 281
				& 159
				& 1801
				& 4890
				& 694
				& 5131
				\\
				& BA & $\bf{1.0000} $ & $\bf{1.0000} $ & $\bf{1.0000} $ & $0.9896 $ & $\bf{1.0000} $ & $\bf{1.0000} $ & $\bf{1.0000} $ & $\bf{1.0000} $ & $\bf{1.0000} $ & $\bf{1.0000} $ & $\bf{1.0000} $ & $\bf{1.0000} $ & $\bf{1.0000} $ & $\bf{1.0000} $\\
				\hline
				\multirow{2}{*}{CSE-BF} & No. attr. ($N3$) 
				& 35
				& 5
				& 5
				& 9
				& 3
				& 12
				& 4
				& 5
				& 5
				& 21
				& 7
				& 14
				& 16
				& 4
				\\
\hline
			\end{tabular}}
			\caption{Balanced accuracy with Random Forest on the train dataset for the classification problems.}
			\label{tab:BATrain}
		\end{center}
	\end{table*}
	
	\begin{table*}[!h]
		\begin{center}
			\resizebox{\textwidth}{!}{\begin{tabular}{llrrrrrrrrrrrrrr}\hline
					\textbf{Method} & \textbf{Metric} 
					& \textbf{amazon-commerce-reviews}
					& \textbf{AP-Colon-Kidney}
					& \textbf{BurkittLymphoma}
					& \textbf{cnae-9}
					& \textbf{Colon}
					& \textbf{eating}
					& \textbf{isolet}
					& \textbf{Lung}
					& \textbf{lymphoma-11classes}
					& \textbf{micro-mass}
					& \textbf{11-tumors}
					& \textbf{GCM}
					& \textbf{tr45.wc}
					& \textbf{rsctc2010-3}
					\\ \hline
					\multirow{2}{*}{All features} & No. attr. ($w$) 
					& $10000$ & $10936$ & $22283$ & $856$ & $2000$ & $6373$ & $617$ & $12600$ & $4026$ & $1300$ & $12533$ & $16063$ & $8261$ & $22277$\\
					& BA & $0.6933 $ & $0.9720 $ & $0.7533 $ & $\bf{1.0000} $ & $\bf{0.8125} $ & $0.5206 $ & $\bf{0.9917} $ & $0.9500 $ & $0.6625 $ & $0.8896 $ & $0.8333 $ & $0.6786 $ & $0.8080 $ & $0.7600 $\\
					\hline
					\multirow{2}{*}{PSEFS-MOEA-V1} & No. attr. ($N1$) 
					& 1018
					& 942
					& 4929
					& 14
					& 20
					& 469
					& 11
					& 1446
					& 69
					& 68
					& 1790
					& 2665
					& 485
					& 5006
					\\
					& BA & $0.7533 $ & $0.9728 $ & $0.8333 $ & $0.9583 $ & $\bf{0.8125} $ & $0.5949 $ & $0.9833 $ & $\bf{1.0000} $ & $0.7500 $ & $0.8808 $ & $0.8636 $ & $0.6667 $ & $0.8376 $ & $0.8000 $\\
					\hline
					\multirow{2}{*}{PSEFS-MOEA-V2} & No. attr. ($N2$) 
					& 1653
					& 994
					& 6560
					& 32
					& 57
					& 1168
					& 14
					& 1626
					& 281
					& 159
					& 1801
					& 4890
					& 694
					& 5131
					\\
					& BA & $\bf{0.7767} $ & $0.9808 $ & $0.8000 $ & $0.9792 $ & $\bf{0.8125} $ & $\bf{0.6004} $ & $0.9833 $ & $\bf{1.0000} $ & $0.7875 $ & $\bf{0.9130} $ & $0.8939 $ & $\bf{0.7262} $ & $0.8453 $ & $0.7600 $\\
					\hline
					\multirow{2}{*}{CSE-BF} & No. attr. ($N3$) 
					& 35
					& 5
					& 5
					& 9
					& 3
					& 12
					& 4
					& 5
					& 5
					& 21
					& 7
					& 14
					& 16
					& 4
					\\
\hline
				\end{tabular}}
				\caption{Balanced accuracy with Random Forest on the test dataset for the classification problems.}
				\label{tab:BATest}
			\end{center}
		\end{table*}
		
\begin{table*}[!h]		
\begin{center}
	\resizebox{\textwidth}{!}{\begin{tabular}{llrrrrrrrrrr}\hline
			\textbf{Method} & \textbf{Metric} 
			& \textbf{liverTox-ALT-target}
			& \textbf{mtp2}
			& \textbf{QSAR-TID-11054}
			& \textbf{QSAR-TID-11110}
			& \textbf{QSAR-TID-11524}
			& \textbf{synthetic-dataset1}
			& \textbf{synthetic-dataset2}
			& \textbf{synthetic-dataset3}
			& \textbf{synthetic-dataset4}
			& \textbf{synthetic-dataset5}
			\\ \hline
			\multirow{2}{*}{All features} & No. attr. ($w$) 
			& $3116$ & $1142$ & $1024$ & $1024$ & $1024$ & $1000$ & $2000$ & $3000$ & $4000$ & $5000$\\
			& RMSE & $517.7284 $ & $0.0545 $ & $0.2258 $ & $0.3256 $ & $0.4106 $ & $481.8693 $ & $664.2083 $ & $800.9976 $ & $980.7938 $ & $1088.4901 $\\
			\hline
			\multirow{2}{*}{PSEFS-MOEA-V1} & No. attr. ($N1$) 
			& 330
			& 183
			& 132
			& 120
			& 176
			& 165
			& 360
			& 571
			& 722
			& 925
			\\
			& RMSE & $528.3113 $ & $0.0529 $ & $0.2337 $ & $0.3765 $ & $0.4260 $ & $469.0402 $ & $652.5475 $ & $779.4592 $ & $949.4602 $ & $1061.0154 $\\
			\hline
			\multirow{2}{*}{PSEFS-MOEA-V2} & No. attr. ($N2$) 
			& 407
			& 240
			& 222
			& 175
			& 187
			& 342
			& 543
			& 822
			& 958
			& 1160
			\\
			& RMSE & $402.3822 $ & $0.0538 $ & $0.2251 $ & $0.3397 $ & $0.4134 $ & $475.5452 $ & $644.3985 $ & $787.9988 $ & $956.1370 $ & $1064.1013 $\\
			\hline
			\multirow{2}{*}{CFSSE-BF} & No. attr. ($N3$) 
			& 14
			& 11
			& 71
			& 77
			& 67
			& 55
			& 76
			& 82
			& 76
			& 63
			\\
\hline
		\end{tabular}}
		\caption{Root mean square error with Random Forest on the train dataset for the regression problems.}
		\label{tab:RMSETrain}
	\end{center}
\end{table*}

\begin{table*}[!h]
	\begin{center}
		\resizebox{\textwidth}{!}{\begin{tabular}{llrrrrrrrrrr}\hline
				\textbf{Method} & \textbf{Metric} 
				& \textbf{liverTox-ALT-target}
				& \textbf{mtp2}
				& \textbf{QSAR-TID-11054}
				& \textbf{QSAR-TID-11110}
				& \textbf{QSAR-TID-11524}
				& \textbf{synthetic-dataset1}
				& \textbf{synthetic-dataset2}
				& \textbf{synthetic-dataset3}
				& \textbf{synthetic-dataset4}
				& \textbf{synthetic-dataset5}
				\\ \hline
				\multirow{2}{*}{All features} & No. attr. ($w$) 
				& $3116$ & $1142$ & $1024$ & $1024$ & $1024$ & $1000$ & $2000$ & $3000$ & $4000$ & $5000$\\
				& RMSE & $3447.5790 $ & $0.1684 $ & $0.8600 $ & $1.1191 $ & $0.8480 $ & $1166.5242 $ & $1747.0308 $ & $2194.1553 $ & $2426.4922 $ & $2700.1656 $\\
				\hline
				\multirow{2}{*}{PSEFS-MOEA-V1} & No. attr. ($N1$) 
				& 330
				& 183
				& 132
				& 120
				& 176
				& 165
				& 360
				& 571
				& 722
				& 925
				\\
				& RMSE & $3126.9837 $ & $\bf{0.1575} $ & $\bf{0.8062} $ & $\bf{1.0124} $ & $0.8441 $ & $1140.9585 $ & $1706.9723 $ & $2171.9144 $ & $\bf{2396.9480} $ & $\bf{2642.1824} $\\
				\hline
				\multirow{2}{*}{PSEFS-MOEA-V2} & No. attr. ($N2$) 
				& 407
				& 240
				& 222
				& 175
				& 187
				& 342
				& 543
				& 822
				& 958
				& 1160
				\\
				& RMSE & $3019.4711 $ & $0.1704 $ & $0.8586 $ & $1.0832 $ & $\bf{0.8221} $ & $1157.5639 $ & $1708.5747 $ & $2182.1657 $ & $2431.6515 $ & $2697.2802 $\\
				\hline
				\multirow{2}{*}{CFSSE-BF} & No. attr. ($N3$) 
				& 14
				& 11
				& 71
				& 77
				& 67
				& 55
				& 76
				& 82
				& 76
				& 63
				\\
\hline
			\end{tabular}}
			\caption{Root mean square error with Random Forest on the test dataset for the regression problems.}
			\label{tab:RMSETest}
		\end{center}
	\end{table*}
	
\begin{table*}[!h]
	\begin{center}
		\resizebox{\textwidth}{!}{\begin{tabular}{llrrrrrrrrrr}\hline
				\textbf{Method} & \textbf{Metric} 
				& \textbf{liverTox-ALT-target}
				& \textbf{mtp2}
				& \textbf{QSAR-TID-11054}
				& \textbf{QSAR-TID-11110}
				& \textbf{QSAR-TID-11524}
				& \textbf{synthetic-dataset1}
				& \textbf{synthetic-dataset2}
				& \textbf{synthetic-dataset3}
				& \textbf{synthetic-dataset4}
				& \textbf{synthetic-dataset5}
				\\ \hline
				\multirow{2}{*}{All features} & No. attr. ($w$) 
				& $3116$ & $1142$ & $1024$ & $1024$ & $1024$ & $1000$ & $2000$ & $3000$ & $4000$ & $5000$\\
				& $R^2$ & $0.9534 $ & $0.9050 $ & $0.9259 $ & $0.9550 $ & $0.8926 $ & $0.8689 $ & $0.8665 $ & $0.8696 $ & $0.8582 $ & $0.8547 $\\
				\hline
				\multirow{2}{*}{PSEFS-MOEA-V1} & No. attr. ($N1$) 
				& 330
				& 183
				& 132
				& 120
				& 176
				& 165
				& 360
				& 571
				& 722
				& 925
				\\
				& $R^2$ & $0.9515 $ & $0.9103 $ & $0.9206 $ & $0.9398 $ & $0.8844 $ & $0.8758 $ & $0.8712 $ & $0.8765 $ & $0.8671 $ & $0.8619 $\\
				\hline
				\multirow{2}{*}{PSEFS-MOEA-V2} & No. attr. ($N2$) 
				& 407
				& 240
				& 222
				& 175
				& 187
				& 342
				& 543
				& 822
				& 958
				& 1160
				\\
				& $R^2$ & $0.9719 $ & $0.9072 $ & $0.9264 $ & $0.9510 $ & $0.8912 $ & $0.8723 $ & $0.8744 $ & $0.8738 $ & $0.8652 $ & $0.8611 $\\
				\hline
				\multirow{2}{*}{CFSSE-BF} & No. attr. ($N3$) 
				& 14
				& 11
				& 71
				& 77
				& 67
				& 55
				& 76
				& 82
				& 76
				& 63
				\\
\hline
			\end{tabular}}
			\caption{$R^2$  with Random Forest on the train dataset for the regression problems.}
			\label{tab:R2Train}
		\end{center}
	\end{table*}
	
	\begin{table*}[!h]
		\begin{center}
			\resizebox{\textwidth}{!}{\begin{tabular}{llrrrrrrrrrr}\hline
					\textbf{Method} & \textbf{Metric} 
					& \textbf{liverTox-ALT-target}
					& \textbf{mtp2}
					& \textbf{QSAR-TID-11054}
					& \textbf{QSAR-TID-11110}
					& \textbf{QSAR-TID-11524}
					& \textbf{synthetic-dataset1}
					& \textbf{synthetic-dataset2}
					& \textbf{synthetic-dataset3}
					& \textbf{synthetic-dataset4}
					& \textbf{synthetic-dataset5}
					\\ \hline
					\multirow{2}{*}{All features} & No. attr. ($w$) 
					& $3116$ & $1142$ & $1024$ & $1024$ & $1024$ & $1000$ & $2000$ & $3000$ & $4000$ & $5000$\\
					& $R^2$ & $0.5838 $ & $0.2208 $ & $0.4296 $ & $0.0699 $ & $0.4863 $ & $0.0167 $ & $-0.0054 $ & $0.0331 $ & $0.0008 $ & $0.0129 $\\
					\hline
					\multirow{2}{*}{PSEFS-MOEA-V1} & No. attr. ($N1$) 
					& 330
					& 183
					& 132
					& 120
					& 176
					& 165
					& 360
					& 571
					& 722
					& 925
					\\
					& $R^2$ & $0.6576 $ & $\bf{0.3183} $ & $\bf{0.4988} $ & $\bf{0.2389} $ & $0.4911 $ & $0.0593 $ & $0.0402 $ & $0.0526 $ & $\bf{0.0250} $ & $\bf{0.0549} $\\
					\hline
					\multirow{2}{*}{PSEFS-MOEA-V2} & No. attr. ($N2$) 
					& 407
					& 240
					& 222
					& 175
					& 187
					& 342
					& 543
					& 822
					& 958
					& 1160
					\\
					& $R^2$ & $0.6808 $ & $0.2021 $ & $0.4316 $ & $0.1286 $ & $\bf{0.5172} $ & $0.0317 $ & $0.0384 $ & $0.0436 $ & $-0.0035 $ & $0.0150 $\\
					\hline
					\multirow{2}{*}{CFSSE-BF} & No. attr. ($N3$) 
					& 14
					& 11
					& 71
					& 77
					& 67
					& 55
					& 76
					& 82
					& 76
					& 63
					\\
\hline
				\end{tabular}}
				\caption{$R^2$  with Random Forest on the test dataset for the regression problems.}
				\label{tab:R2Test}
			\end{center}
		\end{table*}	

\begin{table*}[!h]
\resizebox{\textwidth}{!}{\begin{tabular}{lrrrrrrrrrrrrrr}\hline
		\textbf{Method} 
		& \textbf{amazon-commerce-reviews}
		& \textbf{AP-Colon-Kidney}
		& \textbf{BurkittLymphoma}
		& \textbf{cnae-9}
		& \textbf{Colon}
		& \textbf{eating}
		& \textbf{isolet}
		& \textbf{Lung}
		& \textbf{lymphoma-11classes}
		& \textbf{micro-mass}
		& \textbf{11-tumors}
		& \textbf{GCM}
		& \textbf{tr45.wc}
		& \textbf{rsctc2010-3}
		\\ \hline
		PSEFS-MOEA-V1 & $15521.8000 $ & $5769.0000 $ & $9035.6000 $ & $1190.8000 $ & $1732.4000 $ & $4753.5000 $ & $1091.2000 $ & $5277.9000 $ & $2449.1000 $ & $1861.3000 $ & $5946.1000 $ & $7205.5000 $ & $4775.9000 $ & $9077.5000 $\\
		PSEFS-MOEA-V2 & $44413.8000 $ & $19277.5000 $ & $13666.1000 $ & $1287.1000 $ & $1673.6000 $ & $18407.6000 $ & $1130.5000 $ & $6587.3000 $ & $2885.8000 $ & $2199.8000 $ & $6962.5000 $ & $7205.5000 $ & $15319.7000 $ & $9763.6000 $\\
		CSE-BF & $200.7233 $ & $7.5491 $ & $8.2399 $ & $0.2381 $ & $0.5000 $ & $22.0156 $ & $0.2479 $ & $5.7031 $ & $0.8125 $ & $3.4375 $ & $4.4314 $ & $15.9531 $ & $24.0000 $ & $8.5156 $\\
		CFSSE-BF & $110.3330 $ & $187.6866 $ & $6021.4907 $ & $0.0832 $ & $5.1094 $ & $648.5000 $ & $0.0881 $ & $5424.6406 $ & $137.3438 $ & $6.0312 $ & $2941.0460 $ & $1370.3125 $ & $28.4062 $ & $5583.2500 $\\
		PFI-V1 & $3215.9688 $ & $2496.2500 $ & $8450.4844 $ & $38.1250 $ & $118.9219 $ & $1074.5469 $ & $25.8594 $ & $2844.1406 $ & $358.5938 $ & $76.2031 $ & $2738.2188 $ & $4402.0156 $ & $1607.6875 $ & $7777.0781 $\\
		PFI-V2 & $7038.6250 $ & $3830.7500 $ & $10688.2500 $ & $46.0781 $ & $122.3594 $ & $2077.4844 $ & $33.4375 $ & $3471.8438 $ & $395.9531 $ & $109.0312 $ & $3332.4062 $ & $5436.8125 $ & $2667.6875 $ & $8736.6875 $\\
		SAE & $1.8750 $ & $0.9367 $ & $0.6369 $ & $0.0088 $ & $0.0119 $ & $0.7265 $ & $0.0353 $ & $0.4696 $ & $0.1184 $ & $0.0433 $ & $0.6552 $ & $0.5478 $ & $1.7863 $ & $0.2827 $\\
\hline
	\end{tabular}}
	\caption{Run times (seconds) of the FS methods for the classification problems.}
	\label{tab:RunTimesClassification}
\end{table*}			

\begin{table*}[!h]
\resizebox{\textwidth}{!}{\begin{tabular}{lrrrrrrrrrr}\hline
		\textbf{Method} 
		& \textbf{liverTox-ALT-target}
		& \textbf{mtp2}
		& \textbf{QSAR-TID-11054}
		& \textbf{QSAR-TID-11110}
		& \textbf{QSAR-TID-11524}
		& \textbf{synthetic-dataset1}
		& \textbf{synthetic-dataset2}
		& \textbf{synthetic-dataset3}
		& \textbf{synthetic-dataset4}
		& \textbf{synthetic-dataset5}
		\\ \hline
PSEFS-MOEA-V1 & $1832.8000 $ & $1122.4000 $ & $1119.4000 $ & $1809.4000 $ & $1307.4000 $ & $1743.6000 $ & $2218.4000 $ & $2593.1000 $ & $2987.3000 $ & $3185.3000 $ \\
PSEFS-MOEA-V2 & $1625.5000 $ & $1393.3000 $ & $1386.4000 $ & $2412.1000 $ & $1853.5000 $ & $2806.9000 $ & $3670.1000 $ & $4398.8000 $ & $5137.3000 $ & $5077.5000 $ \\
CFSSE-BF & $3.1945 $ & $0.3647 $ & $4.0899 $ & $4.0654 $ & $3.2705 $ & $2.2524 $ & $15.0988 $ & $39.2173 $ & $52.6996 $ & $61.7243$\\
PFI-V1 & $236.3750 $ & $61.5469 $ & $49.7500 $ & $62.8125 $ & $54.0000 $ & $95.3281 $ & $218.9531 $ & $364.5156 $ & $538.5469 $ & $709.2969 $ \\
PFI-V2 & $252.9688 $ & $78.9375 $ & $65.7812 $ & $111.5000 $ & $84.6875 $ & $160.2812 $ & $374.7031 $ & $641.2812 $ & $925.3906 $ & $1192.3438 $ \\
RFAE & $0.2321 $ & $0.2713 $ & $0.3643 $ & $2.6580 $ & $1.1185 $ & $2.5624 $ & $4.1781 $ & $5.1026 $ & $5.0894 $ & $4.7071 $ \\
CAE & $0.0433 $ & $0.0183 $ & $0.0072 $ & $0.0076 $ & $0.0059 $ & $0.0092 $ & $0.0133 $ & $0.0193 $ & $0.0342 $ & $0.0384 $ \\\hline
	\end{tabular}}
		\caption{Run times (seconds) of the FS methods for the regression problems.}
		\label{tab:RunTimesRegression}
	\end{table*}

\section{Analysis of results and discussion}
\label{sec:analysis}

To rigorously analyze the results obtained in this study, it is essential to employ statistical tests as tools to detect whether the observed differences among the compared FS methods are statistically significant. The use of statistical tests enables a more objective comparison, helping to determine if the observed performance improvements or degradations are not merely due to random chance.

In this work, we compare each FS method against all others and compute the number of times each method statistically outperforms (wins) or underperforms (losses) the others. By calculating the difference between the wins and losses, we can generate a ranking of the methods that highlights their relative effectiveness. This approach provides a clear and interpretable analysis of which feature selection methods are most effective for the given datasets.

Specifically, we have employed the \textit{Wilcoxon signed-rank} test to evaluate the statistical significance of the differences in performance. This non-parametric test is suitable for paired comparisons, where the same datasets are used to evaluate the different FS methods. The test operates under the null hypothesis that there is no difference in the median performance between two methods. By analyzing the ranks of paired differences (ignoring the direction of the difference), the test assesses whether the observed differences are consistent enough to reject the null hypothesis.

In our experiments, the Wilcoxon signed-rank test was applied to the performance results obtained on the test sets for the ACC and BA metrics in the classification problems, and for the RMSE and R2 in the regression problems. A difference between two methods is considered statistically significant if the $ p $-value obtained from the Wilcoxon signed-rank test is less than a predefined significance level, typically 
$\alpha=0.05$. In such cases, we can conclude with confidence that the observed difference in performance is unlikely to have occurred by chance. 

By analyzing the statistical wins-losses across all datasets, we aim to provide a comprehensive and robust evaluation of the proposed method, PSEFS-MOEA, in comparison to the other FS approaches. In addition to comparing the FS methods among themselves, the statistical tests have also included comparisons with models trained using all the features (i.e., without performing feature selection). This additional comparison aims to evaluate whether the application of FS methods leads to improvements in model performance or a reduction in complexity without sacrificing predictive accuracy. By including this baseline, we can assess the value of the proposed and existing FS methods more comprehensively.

The Tables \ref{tab:SignedRank-ACC-test}, \ref{tab:SignedRank-BA-test}, \ref{tab:SignedRank-RMSE-test}, and \ref{tab:SignedRank-R2-test} present the rankings based on statistically significant differences obtained using the Wilcoxon signed-rank test for the metrics ACC, BA, normalized RMSE\footnote{In this statistical analysis, we compare regression model performances across 10 different datasets using the Wilcoxon signed-rank test. Since the output ranges of the datasets vary, using the raw RMSE values could lead to biased conclusions, as the magnitude of RMSE is highly dependent on the scale of the target variable. To ensure a fair comparison, we normalize the RMSE by dividing it by the range of the target variable in each dataset. The normalized RMSE (nRMSE) is calculated as 
$
	nRMSE=
	\frac{RMSE}{\max(y)-\min(y)}
$
where $y$ represents the target variable. Normalizing the RMSE ensures that the metric is independent of the scale of the target variable, allowing for an equitable evaluation of model performance across datasets with different ranges.}, and $R^2$, respectively. Each table summarizes the pairwise comparisons between the FS methods, including the baseline models using all features, highlighting the number of wins and losses for each method. The final rankings are computed based on the difference (wins $-$ losses), allowing for a clearer interpretation of each method's relative performance.

\begin{table}[t!]
	\begin{center}
		\resizebox{0.48\textwidth}{!}{\begin{tabular}{lrrr}
			\hline
			\textbf{Method} & \textbf{Wins} & \textbf{Losses} & \textbf{Wins $-$ Losses} \\\hline
			PSEFS-MOEA-V2 ($N2$ attr.) & 31 & 0 & 31\\
			PSEFS-MOEA-V1 ($N1$ attr.) & 29 & 0 & 29\\
			RFAE ($N2$ attr.) & 18 & 2 & 16\\
			SAE ($N2$ attr.) & 15 & 2 & 13\\
			All features & 14 & 2 & 12\\
			IGAE ($N1$ attr.) & 14 & 2 & 12\\
			IGAE ($N2$ attr.) & 14 & 2 & 12\\
			CSAE ($N2$ attr.) & 13 & 1 & 12\\
			CFSSE-BF ($N4$ attr.) & 13 & 2 & 11\\
			IGAE ($100$ attr.) & 13 & 2 & 11\\
			CSAE ($N1$ attr.) & 12 & 1 & 11\\
			RFAE ($N1$ attr.) & 12 & 2 & 10\\
			CSAE ($100$ attr.) & 12 & 2 & 10\\
			CAE ($100$ attr.) & 12 & 2 & 10\\
			SAE ($100$ attr.) & 12 & 3 & 9\\
			CAE ($N2$ attr.) & 12 & 3 & 9\\
			SAE ($N1$ attr.) & 11 & 3 & 8\\
			CAE ($N1$ attr.) & 11 & 5 & 6\\
			PFI-V1 ($N2$ attr.) & 10 & 5 & 5\\
			RFAE ($100$ attr.) & 9 & 4 & 5\\
			PFI-V1 ($N1$ attr.) & 5 & 9 & -4\\
			PFI-V2 ($N2$ attr.) & 4 & 16 & -12\\
			CSE-BF ($N3$ attr.) & 5 & 18 & -13\\
			PFI-V1 ($100$ attr.) & 3 & 20 & -17\\
			IGAE ($10$ attr.) & 3 & 20 & -17\\
			CAE ($10$ attr.) & 4 & 21 & -17\\
			PFI-V2 ($N1$ attr.) & 1 & 20 & -19\\
			SAE ($10$ attr.) & 1 & 21 & -20\\
			RFAE ($10$ attr.) & 1 & 21 & -20\\
			CSAE ($10$ attr.) & 1 & 23 & -22\\
			PFI-V2 ($100$ attr.) & 1 & 25 & -24\\
			PFI-V1 ($10$ attr.) & 1 & 26 & -25\\
			PFI-V2 ($10$ attr.) & 0 & 32 & -32\\
			\hline
		\end{tabular}
	}
		\caption{Ranking of statistically significant differences using the Wilcoxon signed-rank test for classification problems with the ACC metric.}
		\label{tab:SignedRank-ACC-test}
	\end{center}
\end{table}

\begin{table}[t!]
	\begin{center}
		\resizebox{0.48\textwidth}{!}{\begin{tabular}{lrrr}
			\hline
			\textbf{Method} & \textbf{Wins} & \textbf{Losses} & \textbf{Wins $-$ Losses} \\\hline
			PSEFS-MOEA-V2 ($N2$ attr.) & 25 & 0 & 25\\
			PSEFS-MOEA-V1 ($N1$ attr.) & 23 & 0 & 23\\
			RFAE ($N2$ attr.) & 17 & 2 & 15\\
			IGAE ($N1$ attr.) & 14 & 0 & 14\\
			SAE ($N2$ attr.) & 15 & 2 & 13\\
			CFSSE-BF ($N4$ attr.) & 13 & 1 & 12\\
			SAE ($100$ attr.) & 12 & 0 & 12\\
			IGAE ($100$ attr.) & 12 & 0 & 12\\
			IGAE ($N2$ attr.) & 14 & 2 & 12\\
			RFAE ($N1$ attr.) & 13 & 1 & 12\\
			CSAE ($100$ attr.) & 12 & 0 & 12\\
			CSAE ($N1$ attr.) & 12 & 0 & 12\\
			CSAE ($N2$ attr.) & 13 & 1 & 12\\
			All features & 13 & 2 & 11\\
			CAE ($100$ attr.) & 11 & 1 & 10\\
			CAE ($N2$ attr.) & 13 & 3 & 10\\
			SAE ($N1$ attr.) & 12 & 3 & 9\\
			RFAE ($100$ attr.) & 10 & 3 & 7\\
			CAE ($N1$ attr.) & 9 & 5 & 4\\
			PFI-V1 ($N2$ attr.) & 6 & 4 & 2\\
			PFI-V1 ($N1$ attr.) & 3 & 12 & -9\\
			PFI-V2 ($N2$ attr.) & 4 & 16 & -12\\
			CSE-BF ($N3$ attr.) & 3 & 16 & -13\\
			SAE ($10$ attr.) & 3 & 18 & -15\\
			CAE ($10$ attr.) & 4 & 19 & -15\\
			IGAE ($10$ attr.) & 3 & 19 & -16\\
			PFI-V1 ($100$ attr.) & 3 & 20 & -17\\
			PFI-V2 ($N1$ attr.) & 2 & 21 & -19\\
			RFAE ($10$ attr.) & 1 & 20 & -19\\
			CSAE ($10$ attr.) & 1 & 20 & -19\\
			PFI-V2 ($100$ attr.) & 1 & 27 & -26\\
			PFI-V1 ($10$ attr.) & 1 & 28 & -27\\
			PFI-V2 ($10$ attr.) & 0 & 32 & -32\\
			\hline
		\end{tabular}
	}
		\caption{Ranking of statistically significant differences using the Wilcoxon signed-rank test for classification problems with the BA metric.}
		\label{tab:SignedRank-BA-test}
	\end{center}
\end{table}

\begin{table}[t!]
	\begin{center}
		\resizebox{0.48\textwidth}{!}{\begin{tabular}{lrrr}
			\hline
			\textbf{Method} & \textbf{Wins} & \textbf{Losses} & \textbf{Wins $-$ Losses} \\\hline
PSEFS-MOEA-V1 ($N1$ attr.) & 15 & 0 & 15\\
PSEFS-MOEA-V2 ($N2$ attr.) & 8 & 0 & 8\\
PFI-V2 ($100$ attr.) & 6 & 1 & 5\\
CAE ($100$ attr.) & 2 & 0 & 2\\
CFSSE-BF ($N3$ attr.) & 1 & 0 & 1\\
PFI-V2 ($10$ attr.) & 1 & 0 & 1\\
All features & 1 & 1 & 0\\
PFI-V1 ($N2$ attr.) & 1 & 1 & 0\\
CAE ($N1$ attr.) & 1 & 1 & 0\\
CAE ($N2$ attr.) & 1 & 1 & 0\\
PFI-V1 ($10$ attr.) & 0 & 1 & -1\\
PFI-V1 ($100$ attr.) & 1 & 2 & -1\\
PFI-V1 ($N1$ attr.) & 1 & 2 & -1\\
PFI-V2 ($N1$ attr.) & 1 & 2 & -1\\
PFI-V2 ($N2$ attr.) & 1 & 2 & -1\\
RFAE ($100$ attr.) & 1 & 3 & -2\\
RFAE ($N1$ attr.) & 1 & 3 & -2\\
RFAE ($N2$ attr.) & 1 & 3 & -2\\
CAE ($10$ attr.) & 0 & 4 & -4\\
RFAE ($10$ attr.) & 0 & 17 & -17\\
			\hline
		\end{tabular}
	}
		\caption{Ranking of statistically significant differences using the Wilcoxon signed rank test for regression problems with the nRMSE metric}
		\label{tab:SignedRank-RMSE-test}
	\end{center}
\end{table}

\begin{table}[t!]
	\begin{center}
		\resizebox{0.48\textwidth}{!}{\begin{tabular}{lrrr}
			\hline
			\textbf{Method} & \textbf{Wins} & \textbf{Losses} & \textbf{Wins $-$ Losses} \\\hline
			PSEFS-MOEA-V1 ($N1$ attr.) & 17 & 0 & 17\\
			PSEFS-MOEA-V2 ($N2$ attr.) & 10 & 1 & 9\\
			PFI-V2 ($100$ attr.) & 8 & 1 & 7\\
			CAE ($100$ attr.) & 4 & 0 & 4\\
			CAE ($N1$ attr.) & 4 & 1 & 3\\
			CAE ($N2$ attr.) & 4 & 1 & 3\\
			PFI-V1 ($100$ attr.) & 4 & 2 & 2\\
			PFI-V1 ($N1$ attr.) & 4 & 2 & 2\\
			PFI-V2 ($N1$ attr.) & 4 & 2 & 2\\
			PFI-V2 ($N2$ attr.) & 4 & 2 & 2\\
			All features & 2 & 1 & 1\\
			CFSSE-BF ($N3$ attr.) & 1 & 0 & 1\\
			PFI-V1 ($N2$ attr.) & 2 & 1 & 1\\
			RFAE ($N2$ attr.) & 4 & 3 & 1\\
			RFAE ($100$ attr.) & 1 & 3 & -2\\
			RFAE ($N1$ attr.) & 1 & 3 & -2\\
			PFI-V2 ($10$ attr.) & 1 & 11 & -10\\
			CAE ($10$ attr.) & 0 & 11 & -11\\
			PFI-V1 ($10$ attr.) & 0 & 13 & -13\\
			RFAE ($10$ attr.) & 0 & 17 & -17\\
			\hline
		\end{tabular}
	}
		\caption{Ranking of statistically significant differences using the signed rank test for regression problems with the $R^2$ metric}
		\label{tab:SignedRank-R2-test}
	\end{center}
\end{table}

To further analyze the detailed comparisons between the two proposed versions of PSEFS-MOEA (V1 and V2), the comparisons of PSEFS-MOEA with the conventional PFI method (also in its two versions, V1 and V2), and the comparisons of PSEFS-MOEA with models built using all features (i.e., without applying feature selection), Table \ref{tab:p-values} presents the $ p $-values obtained from statistical tests for these specific pairwise comparisons. These include: PSEFS-MOEA-V1 vs. PSEFS-MOEA-V2, PSEFS-MOEA-V1 vs. PFI-V1 (with 10, 100, $ N1 $, and $ N2 $ features), PSEFS-MOEA-V1 vs. PFI-V2 (with 10, 100, $ N1 $, and $ N2 $ features), PSEFS-MOEA-V2 vs. PFI-V1 (with 10, 100, $ N1 $, and $ N2 $ features), PSEFS-MOEA-V2 vs. PFI-V2 (with 10, 100, $ N1 $, and $ N2 $ features), PSEFS-MOEA-V1 vs. models using all features, and PSEFS-MOEA-V2 vs. models using all features.

For a thorough assessment, when the $ p $-value exceeds the threshold of 0.05 --indicating no statistically significant difference between the two compared methods-- we also computed the difference in mean metric values for each pair of compared methods (mean of Method 1 minus mean of Method 2 for ACC, BA and $R^2$, and mean of Method 2 minus mean of Method 1 for nRMSE). These differences, shown in parentheses in Table \ref{tab:p-values}, allow us to discern trends in performance. A positive difference indicates that Method 1 outperforms Method 2 on average for a specific metric, whereas a negative difference suggests the opposite.
In Table \ref{tab:p-values}, $ p $-values and mean differences that highlight Method 1 as superior to Method 2 (either statistically or on average) are emphasized in bold for clarity. This additional analysis enables a more nuanced interpretation of the results, particularly in cases where statistical significance is not observed.

\begin{table*}[!t]
 	\centering
 	\resizebox{\textwidth}{!}{
 		\begin{tabular}{llllll}
 			\hline
 			\textbf{Method 1} & \textbf{Method 2} & \textbf{ACC} & \textbf{BA} & \textbf{nRMSE} & $\bf{\textit{R}^2}$ \\
 			\hline
 			PSEFS-MOEA-V1 ($N1$ attr.) & PSEFS-MOEA-V2 ($N2$ attr.) & $0.1849$ ($-0.0071$) & $0.1823$ ($-0.0108$) & $0.1602$ ($\bf{0.0031}$) &  $\bf{0.0371}$ \\\hline
 			
 			\multirow{9}{*}{PSEFS-MOEA-V1 ($N1$ attr.)}
 			& PFI-V1 (10 attr.)    & $\bf{0.0019}$ & $\bf{0.0019}$ & $\bf{0.0488}$  & $\bf{0.0098}$ \\
 			& PFI-V1 (100 attr.)   & $\bf{0.0022}$ & $\bf{0.0022}$ & $\bf{0.0059}$  & $\bf{0.0039}$ \\
 			& PFI-V1 ($N1$ attr.)  & $\bf{0.0029}$ & $\bf{0.0029}$ & $\bf{0.0020}$  & $\bf{0.0020}$ \\
 			& PFI-V1 ($N2$ attr.)  & $\bf{0.0044}$ & $\bf{0.0044}$ & $\bf{0.0020}$  & $\bf{0.0020}$ \\\cline{2-6}
 			& PFI-V2 (10 attr.)    & $\bf{0.0002}$ & $\bf{0.0002}$ & $0.0645$ ($\bf{0.0104}$)      & $\bf{0.0059}$ \\
 			& PFI-V2 (100 attr.)   & $\bf{0.0019}$ & $\bf{0.0019}$ & $\bf{0.0371}$ & $\bf{0.0195}$ \\
 			& PFI-V2 ($N1$ attr.)  & $\bf{0.0004}$ & $\bf{0.0004}$ & $\bf{0.0039}$ & $\bf{0.0039}$ \\
 			& PFI-V2 ($N2$ attr.)  & $\bf{0.0024}$ & $\bf{0.0024}$ & $\bf{0.0039}$ & $\bf{0.0039}$ \\\hline
 			
 			\multirow{9}{*}{PSEFS-MOEA-V2 ($N2$ attr.)}
 			& PFI-V1 (10 attr.)    & $\bf{0.0002}$ & $\bf{0.0006}$ & $0.0645$ ($\bf{0.0120}$)      & $\bf{0.0098}$ \\
 			& PFI-V1 (100 attr.)   & $\bf{0.0015}$ & $\bf{0.0015}$ & $\bf{0.0273}$ & $\bf{0.0273}$ \\
 			& PFI-V1 ($N1$ attr.)  & $\bf{0.0019}$ & $\bf{0.0024}$ & $\bf{0.0371}$ & $\bf{0.0195}$ \\
 			& PFI-V1 ($N2$ attr.)  & $\bf{0.0029}$ & $\bf{0.0029}$ & $0.1602$ ($\bf{0.0036}$)     & $0.1055$ ($\textbf{0.0196}$) \\\cline{2-6}
 			& PFI-V2 (10 attr.)    & $\bf{0.0015}$ & $\bf{0.0015}$ & $0.0645$ ($\bf{0.0073}$)     & $\bf{0.0137}$ \\
 			& PFI-V2 (100 attr.)   & $\bf{0.0022}$ & $\bf{0.0022}$ & $0.6250$ ($\bf{0.0015}$)     & $1.0000$ ($\bf{0.0085}$) \\
 			& PFI-V2 ($N1$ attr.)  & $\bf{0.0022}$ & $\bf{0.0022}$ & $0.0645$ ($\bf{0.0031}$)     & $0.0645$ ($\bf{0.0243}$) \\
 			& PFI-V2 ($N2$ attr.)  & $\bf{0.0051}$ & $\bf{0.0051}$ & $\bf{0.0488}$ & $\bf{0.0273}$ \\
 			\hline
 			
 			PSEFS-MOEA-V1 ($N1$ attr.) & All features &  $\bf{0.0469}$ & $\bf{0.0464}$ & $\bf{0.0020}$ & $\bf{0.0020}$ \\
 			PSEFS-MOEA-V2 ($N2$ attr.) & All features &  $\bf{0.0076}$ & $\bf{0.0060}$ & $0.0645$  ($\bf{0.0041}$)    & $0.0645$ ($\bf{0.0237}$) \\\hline
 	\end{tabular}
}
\caption{$p$-values of the comparisons between the two versions of PSEFS-MOEA, and the comparisons of PSEFS-MOEA  with PFI and the models built with all features. Mean differences are shown in brackets when there are no statistically significant differences ($p$-value $\geq 0.05$). A positive difference indicates that Method 1 outperforms Method 2 on average for a specific metric, whereas a negative difference suggests the opposite. $ p $-values and mean differences that highlight Method 1 as superior to Method 2 (either statistically or on average) are emphasized in bold.}
\label{tab:p-values}
\end{table*}

The analysis of the results obtained, based on the tables presented and the statistical tests discussed above, is presented below. This analysis is organised in accordance with the specific objectives outlined in section \ref{sec:descexp}.

\subsection{Comparison of PSEFS-MOEA versions}
The comparison between PSEFS-MOEA-V1 and PSEFS-MOEA-V2 reveals no statistically significant differences for ACC, BA, or nRMSE ($ p $-value $\geq 0.05$ in all cases). However, for $R^2$, the 
$p$-value (0.0371) indicates that PSEFS-MOEA-V1 achieves statistically significant improvements over V2. To further interpret the results, mean differences are examined for metrics without statistical significance:

\begin{itemize}
	\item ACC and BA: Both differences ($-0.0071$ and $-0.0108$, respectively) are slightly negative, suggesting V1 might underperform V2 in these metrics, though the differences are negligible.
	\item nRMSE: A small positive difference (+0.0031) hints at better performance for V1 in terms of error minimization, aligning with the observed $R^2$ improvements.
\end{itemize}

In summary, although the differences are subtle in most metrics, V2 shows a slight advantage in classification tasks, while in regression tasks V1 shows statistical differences over V2 in the $R^2$ metric and a slight advantage in the nRMSE metric.

\subsection{Evaluation against conventional PFI} 
For classification tasks, PSEFS-MOEA-V1 and PSEFS-MOEA-V2 consistently outperform PFI with statistically significant results ($p$-value $< 0.05$) for both metrics ACC and BA.
This demonstrates the robustness of PSEFS-MOEA compared to PFI, regardless of the number of selected attributes or metric evaluated.
For regression tasks, PSEFS-MOEA-V1 and PSEFS-MOEA-V2 shows statistically significant superiority ($p$-value $< 0.05$)) over PFI for most comparisons. In certain configurations 
$p$-value $\geq 0.05$. However, mean differences suggest that PSEFS-MOEA-V1 and PSEFS-MOEA-V2 still outperforms PFI on average for these configurations.

Overall, PSEFS-MOEA demonstrates clear advantages over PFI in both classification and regression tasks, regardless of the metric or configuration considered.

\subsection{Effectiveness in reducing feature sets}
When comparing PSEFS-MOEA-V1 to models with all features, statistically significant improvements ($p$-values $<0.05$) are observed for all metrics, including ACC, BA, nRMSE and 
$R^2$. This highlights that the feature selection performed by V1 enhances predictive performance while reducing the number of features used. Similar trends are observed when comparing the PSEFS-MOEA-V2 with the models with all features, but for the nRMSE and $R^2$, the $p$-values are $>0.05$. Mean differences (0.0041 for nRMSE and 0.0237 for $R^2$) indicate that V2 still outperforms models with all features on average.

In conclusion, both versions of PSEFS-MOEA achieve better predictive performance than models with all features, while simultaneously reducing feature sets, with V1 achieving consistent statistical significance and V2 showing average improvements.

\subsection{Comparison with established methods} The results of the ranking of statistically significant differences between the methods evaluated (including PSEFS-MOEA-V1, PSEFS-MOEA-V2, and other established methods) confirm the robustness of the proposed approaches across the four evaluation metrics (ACC, BA, nRMSE, and 
$R^2$). 

For the classification tasks (Tables \ref{tab:SignedRank-ACC-test} and \ref{tab:SignedRank-BA-test}):

\begin{itemize}
	\item PSEFS-MOEA-V2 achieved the highest position in the ranking for both ACC and BA metrics, with wins-losses differences of +31 and +25, respectively. This confirms its strong performance in classification tasks.
	\item PSEFS-MOEA-V1 ranked second in both metrics (ACC: +29, BA: +23), consistently outperforming all other methods except its successor, PSEFS-MOEA-V2.
	\item The performance gap between PSEFS-MOEA-V2 and V1 highlights that V2 is slightly more effective than V1 in classification problems. This aligns with the earlier analysis of classification-specific comparisons between these two versions.
\end{itemize}

For the regression tasks (Tables \ref{tab:SignedRank-RMSE-test} and \ref{tab:SignedRank-R2-test}):

\begin{itemize}
	\item PSEFS-MOEA-V1 emerged as the top performer in regression tasks for both nRMSE (+15) and $R^2$ (+17). This confirms its capability to minimize error and improve predictive accuracy in regression problems.
	\item PSEFS-MOEA-V2 followed as the second-best method, with a wins-losses difference of +8 for nRMSE and +9 for $R^2$.
	\item The rankings show a reversal in performance trends compared to classification tasks, with PSEFS-MOEA-V1 outperforming V2 in regression scenarios. This observation corroborates the earlier findings of superior nRMSE and $R^2$ performance by PSEFS-MOEA-V1.
\end{itemize}

In general, both versions of PSEFS-MOEA dominate the rankings across all metrics, consistently outperforming established subset evaluation filter methods such as CSE-BF and CFSSE-BF, attribute evaluation filter methods such as IGAE, RFAE, SAE, PFI, CSAE and CAE, and models built with all features. The ranking analysis reinforces the adaptability and effectiveness of PSEFS-MOEA versions, with PSEFS-MOEA-V2 excelling in classification tasks and PSEFS-MOEA-V1 leading in regression scenarios. 

\subsubsection*{Overfitting analysis}
To further support the performance analysis, Tables \ref{tab:overfitting-ACC-BA} and \ref{tab:overfitting-RMSE-R2} present the degrees of overfitting for the models obtained with all FS methods and the models built using all features.

\begin{table}[!t]
	\begin{center}
		\begin{tabular}{lccc}\hline
			\textbf{Method} & \textbf{No. attr.} & $\bf{\frac{ACC_{train}}{ACC_{test}}}$ & $\bf{\frac{BA_{train}}{BA_{test}}}$\\ 
			\hline
			All features & $w$  & $1.2131$ & $1.2766$ \\ 
			\hline 
			PSEFS-MOEA-V1  & $N1$ & $1.1637$ & $1.2186$ \\ 
			\hline 
			PSEFS-MOEA-V2  & $N2$ & $1.1541$ & $1.2029$ \\ 
			\hline 
			CSE-BF & $N3$ & $1.4046$ & $1.4789$ \\ 
			\hline 
			CFSSE-BF & $N4$ & $1.2063$ & $1.2597$ \\ 
			\hline 
			\multirow{4}{*}{PFI-V1} & $10$ & $1.8774$ & $2.2329$ \\ 
			& $100$ & $1.3665$ & $1.5024$ \\ 
			& $N1$ & $1.3206$ & $1.4378$ \\ 
			& $N2$ & $1.2626$ & $1.3648$ \\ 
			\hline
			\multirow{4}{*}{PFI-V2} & $10$ & $2.3733$ & $2.8169$ \\ 
			& $100$ & $1.8304$ & $2.0495$ \\ 
			& $N1$ & $1.5767$ & $1.6736$ \\ 
			& $N2$ & $1.3794$ & $1.4796$ \\ 
			\hline
			\multirow{4}{*}{SAE} & $10$ & $1.4566$ & $1.5301$ \\ 
			& $100$ & $1.2614$ & $1.2956$ \\ 
			& $N1$ & $1.2536$ & $1.3143$ \\ 
			& $N2$ & $1.2007$ & $1.2604$ \\ 
			\hline
			\multirow{4}{*}{IGAE} & $10$ & $1.6064$ & $1.6835$ \\ 
			& $100$ & $1.2491$ & $1.3279$ \\ 
			& $N1$ & $1.2162$ & $1.2667$ \\ 
			& $N2$ & $1.2072$ & $1.2731$ \\ 
			\hline
			\multirow{4}{*}{RFAE} & $10$ & $1.7088$ & $1.8329$ \\ 
			& $100$ & $1.3314$ & $1.4490$ \\ 
			& $N1$ & $1.2502$ & $1.3056$ \\ 
			& $N2$ & $1.2094$ & $1.2702$ \\ 
			\hline
			\multirow{4}{*}{CSAE} & $10$ & $1.5548$ & $1.6178$ \\ 
			& $100$ & $1.2657$ & $1.3125$ \\ 
			& $N1$ & $1.2243$ & $1.2753$ \\ 
			& $N2$ & $1.2115$ & $1.2700$ \\ 
			\hline
			\multirow{4}{*}{CAE} & $10$ & $1.5288$ & $1.6182$ \\ 
			& $100$ & $1.2460$ & $1.3093$ \\ 
			& $N1$ & $1.2476$ & $1.3284$ \\ 
			& $N2$ & $1.2244$ & $1.2888$ \\ 
			\hline
		\end{tabular}
		\caption{Overfitting ratios for the classification tasks.}
		\label{tab:overfitting-ACC-BA}
	\end{center}
\end{table}

\begin{table}[!t]
	\begin{center}
		\begin{tabular}{lccc}\hline
			\textbf{Method} & \textbf{No. attr.} & $\bf{\frac{{RMSE}_{test}}{{RMSE}_{train}}}$ & $\bf{\textit{R}^2_{train}-\textit{R}^2_{test}}$\\ 
			\hline
			All attributes & $w$  & $3.1807$ & $0.7101$ \\ 
			\hline 
			PSEFS-MOEA-V1  & $N1$ & $2.9864$ & $0.6523$ \\ 
			\hline 
			PSEFS-MOEA-V2  & $N2$ & $3.2594$ & $0.6909$ \\ 
			\hline 
			CFSSE-BF & $N3$ & $3.1021$ & $0.6910$ \\ 
			\hline 
			\multirow{4}{*}{PFI-V1} & $10$ & $2.7385$ & $0.7124$ \\ 
			& $100$ & $3.1384$ & $0.7175$ \\ 
			& $N1$ & $3.0545$ & $0.7086$ \\ 
			& $N2$ & $3.1861$ & $0.7083$ \\ 
			\hline
			\multirow{4}{*}{PFI-V2} & $10$ & $3.2369$ & $0.7057$ \\ 
			& $100$ & $3.5099$ & $0.7069$ \\ 
			& $N1$ & $3.3106$ & $0.7170$ \\ 
			& $N2$ & $3.3595$ & $0.7161$ \\ 
			\hline
			\multirow{4}{*}{RFAE} & $10$ & $2.9064$ & $0.7767$ \\ 
			& $100$ & $3.1849$ & $0.7372$ \\ 
			& $N1$ & $3.2697$ & $0.7392$ \\ 
			& $N2$ & $3.2094$ & $0.7198$ \\ 
			\hline
			\multirow{4}{*}{CAE} & $10$ & $2.9667$ & $0.6849$ \\ 
			& $100$ & $3.1020$ & $0.6892$ \\ 
			& $N1$ & $3.0936$ & $0.6877$ \\ 
			& $N2$ & $3.1313$ & $0.6925$ \\ 
			\hline
		\end{tabular}
		\caption{Degrees of overfitting fpor the regression tasks}
		\label{tab:overfitting-RMSE-R2}
	\end{center}
\end{table}

For the classification metrics (ACC and BA), the overfitting ratios were calculated as $\frac{\text{ACC}_{\text{train}}}{\text{ACC}_{\text{test}}}$ and 
$\frac{\text{BA}_{\text{train}}}{\text{BA}_{\text{test}}}$. For the regression metric nRMSE, the overfitting ratio was computed as 
$\frac{\text{nRMSE}_{\text{test}}}{\text{nRMSE}_{\text{train}}}$.
For $R^2$, overfitting was determined as the difference $R^2_{\text{train}}-R^2_{\text{test}}$. Ratios greater than 1 for ACC, BA and nRMSE indicate overfitting, with higher values denoting more severe overfitting. Positive differences for $R^2$ indicate overfitting, with larger differences indicating greater overfitting.

For the classification tasks (Table \ref{tab:overfitting-ACC-BA}), PSEFS-MOEA-V1 and PSEFS-MOEA-V2 exhibited the smallest overfitting ratios for both ACC and BA metrics.These results highlight the effectiveness of the proposed methods in mitigating overfitting compared to all other FS methods and models trained with all features. The low overfitting ratios emphasize the generalization ability of PSEFS-MOEA methods, making them highly reliable for classification tasks.

For the regression tasks (Table \ref{tab:overfitting-RMSE-R2}), PSEFS-MOEA-V1 achieved the top rank with the smallest overfitting difference among all methods for $R^2$ metric.
For nRMSE, PSEFS-MOEA-V1 ranked 4th out of 20, further showcasing its ability to minimize overfitting in regression tasks.
PSEFS-MOEA-V2, while performing well overall, ranked 16th for nRMSE and 5th for $R^2$, indicating a slightly higher tendency for overfitting in regression problems compared to PSEFS-MOEA-V1.
In general, the overfitting analysis confirms the superiority of PSEFS-MOEA-V1 in regression tasks, aligning with its top performance rankings for nRMSE and  $R^2$
metrics.

This comprehensive evaluation underscores the robustness of the PSEFS-MOEA methods, particularly their ability to achieve competitive performance while maintaining low degrees of overfitting, which is critical for real-world applications.

\subsubsection*{Execution times}

The execution times for the PSEFS-MOEA methods, both V1 and V2, were the highest among all the compared methods for classification and regression tasks, as shown in Tables \ref{tab:Runtime-Classification} and \ref{tab:Runtime-Regression}, respectively. The extended computational cost of PSEFS-MOEA methods is an expected consequence of their multi-objective, population-based, meta-heuristic nature. Key factors contributing to these higher runtimes include:

\begin{itemize}
\item \textit{Population-based search}:
Unlike single-solution optimization techniques, MOEAs operate on a population of candidate solutions, iteratively evaluating and evolving this population. This increases the computational effort due to multiple fitness evaluations per iteration.

\item \textit{Multi-objective optimization}:
PSEFS-MOEA methods are designed to optimize multiple conflicting objectives (e.g., feature reduction and predictive performance), which inherently demands more complex calculations and trade-off management than single-objective methods.

\item \textit{Diversity preservation mechanisms}:
MOEAs employ mechanisms such as crowding distance or Pareto dominance sorting to maintain diversity in the solution set, ensuring robust exploration of the search space. These processes add computational overhead compared to simpler FS methods.

\item \textit{Evaluation of features}:
Both versions of PSEFS-MOEA evaluate subsets of features iteratively across a large number of candidate solutions, requiring repeated evaluation of samples in machine learning models, which significantly increases runtime.
\end{itemize}

\begin{table}[!h]
	\centering
\begin{tabular}{lrr}\hline
			\textbf{Method} 
			& \textbf{Average} & \textbf{Ranking}
			\\ \hline
			PSEFS-MOEA-V1 & $5406.2571$ & $10$\\
			PSEFS-MOEA-V2 & $10770.0286$ & $11$\\
			CSE-BF & $21.5977$ & $6$\\
			CFSSE-BF & $1604.5944$ & $7$\\
			PFI-V1 & $2516.0067$ & $8$\\
			PFI-V2 & $3427.6719$ & $9$\\
			SAE & $0.5810$ & $4$\\
			IGAE & $0.5024$ & $3$\\
			RFAE & $11.8244$ & $5$\\
			CSAE & $0.4437$ & $2$\\
			CAE & $0.2332$ & $1$\\
			\hline
		\end{tabular}
		\caption{Average run times (seconds) and ranking of the FS methods for the classification problems.}
		\label{tab:Runtime-Classification}
	\end{table}

\begin{table}[!h]
	\centering
	\begin{tabular}{lrr}\hline
			\textbf{Method} 
			& \textbf{Average} & \textbf{Ranking}
			\\ \hline
			PSEFS-MOEA-V1 & $1991.9100$ & $6$\\
			PSEFS-MOEA-V2 & $2976.1400$ & $7$\\
			BF-CFSSE & $18.5977$ & $3$\\
			PFI-V1 & $239.1125$ & $4$\\
			PFI-V2 & $388.7875$ & $5$\\
			RFAE & $2.6284$ & $2$\\
			CAE & $0.0197$ & $1$\\
			\hline
		\end{tabular}
		\caption{Average run times (seconds) and ranking of the FS methods for the regression problems.}
		\label{tab:Runtime-Regression}
	\end{table}

The sophisticated design of the PSEFS-MOEA methods focuses on achieving high-quality solutions, often prioritizing solution accuracy and robustness over computational efficiency.
While the longer execution times may be seen as a disadvantage, they are justified by the superior performance and generalization capabilities demonstrated by PSEFS-MOEA methods:

\begin{itemize}
	\item As shown in the analysis of classification and regression tasks, the two versions consistently achieved top positions in performance rankings, effectively balancing feature selection and predictive accuracy.

	\item Their ability to reduce overfitting and produce robust models makes them particularly suitable for high-stakes applications where quality outweighs computational cost.
\end{itemize}

As a practical consideration, the additional runtime required by PSEFS-MOEA methods can be mitigated in real-world scenarios by parallel processing or leveraging high-performance computing environments. For applications where computational resources or time are constrained, the methods' efficiency could be further improved by exploring strategies such as reducing population size, iterations, or adopting surrogate models \cite{ESPINOSA2024101587}.
In summary, while the execution times are higher, they are an inherent and acceptable trade-off for the quality, robustness, and reliability of the solutions produced by the PSEFS-MOEA methods.

\subsection{Robustness across tasks}
The proposed PSEFS-MOEA method, in both its V1 and V2 versions, has demonstrated remarkable robustness across a wide variety of tasks, encompassing both classification and regression problems.

\paragraph{Classification tasks} The results indicate consistent and strong performance for both binary and multi-class classification problems. This robustness highlights the method's adaptability to varying levels of class complexity. Among the classification tasks tested, the majority (11 out of 14) were imbalanced datasets, which are particularly challenging due to the potential for bias towards the majority class.
The strong performance of PSEFS-MOEA, particularly in terms of the BA metric, underscores its capability to handle imbalanced problems effectively. This is crucial in applications where the minority class carries significant importance, such as in medical diagnosis or fraud detection.

\paragraph{Regression tasks}
The regression tasks included synthetic problems designed to progressively increase in complexity, with challenges such as higher dimensionality and the addition of noise.
PSEFS-MOEA proved robust in these scenarios, maintaining competitive performance as problem difficulty escalated. This demonstrates the method's ability to generalize well, even under challenging conditions. The robustness against noisy data and high-dimensional spaces reflects the strength of PSEFS-MOEA in effectively selecting relevant features while minimizing overfitting, as discussed earlier.

\paragraph{Hypervolume and Pareto fronts}
To complete the results analysis and demonstrate the robustness of PSEFS-MOEA-V1 and PSEFS-MOEA-V2, Figures \ref{fig:isolet}, \ref{fig:micro-mass}, and \ref{fig:QSAR-TID-11054} illustrate the \textit{hypervolume} \cite{Zit02} evolution of both algorithms and the obtained Pareto fronts for a representative set of problems, given the extensive number of datasets tested in this paper. This representative set includes a binary balanced classification problem (isolet dataset, Figure \ref{fig:isolet}), a multi-class imbalanced classification problem (micro-mass dataset, Figure \ref{fig:micro-mass}), and a regression problem (QSAR-TID-11054 dataset, Figure \ref{fig:QSAR-TID-11054}). The graphs highlight the proper hypervolume evolution of both PSEFS-MOEA-V1 and PSEFS-MOEA-V2, indicating a good balance between convergence and diversity, as further reflected in the Pareto fronts.

\begin{table*}[h]
	\centering
	\resizebox{\textwidth}{!}{
		\begin{tabular}{cc}
			
			\includegraphics[width=1\textwidth]{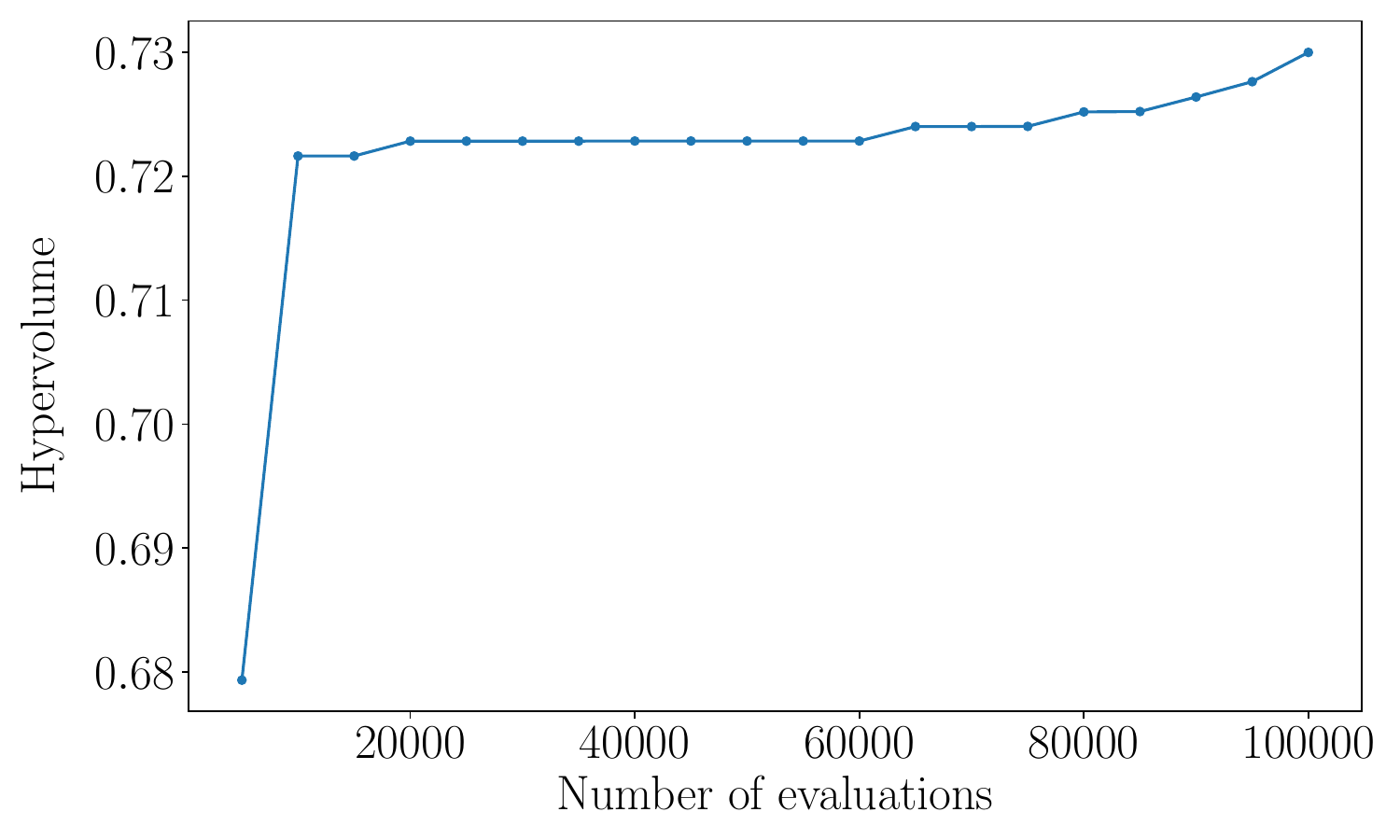} & \includegraphics[width=1\textwidth]{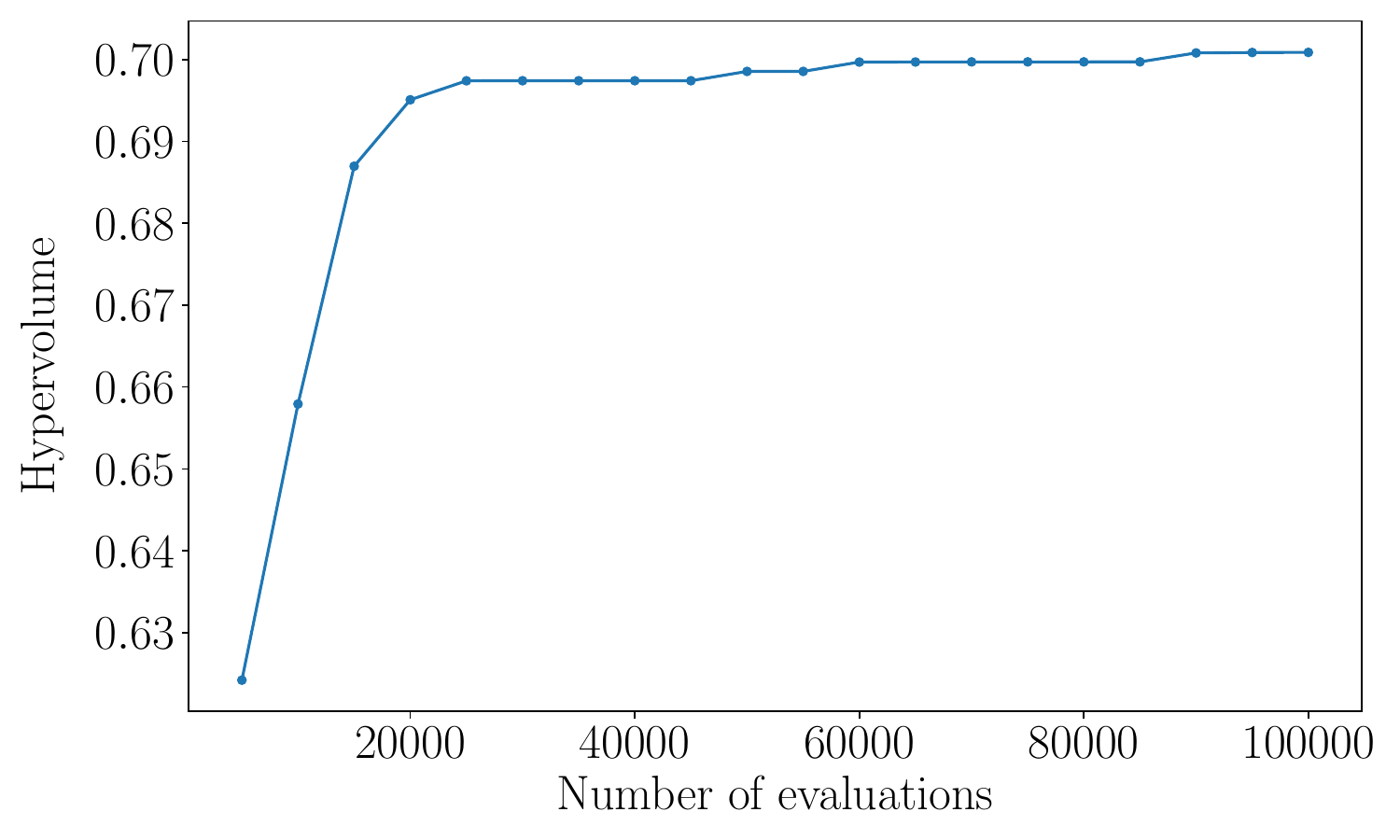}\\
			\includegraphics[width=1\textwidth]{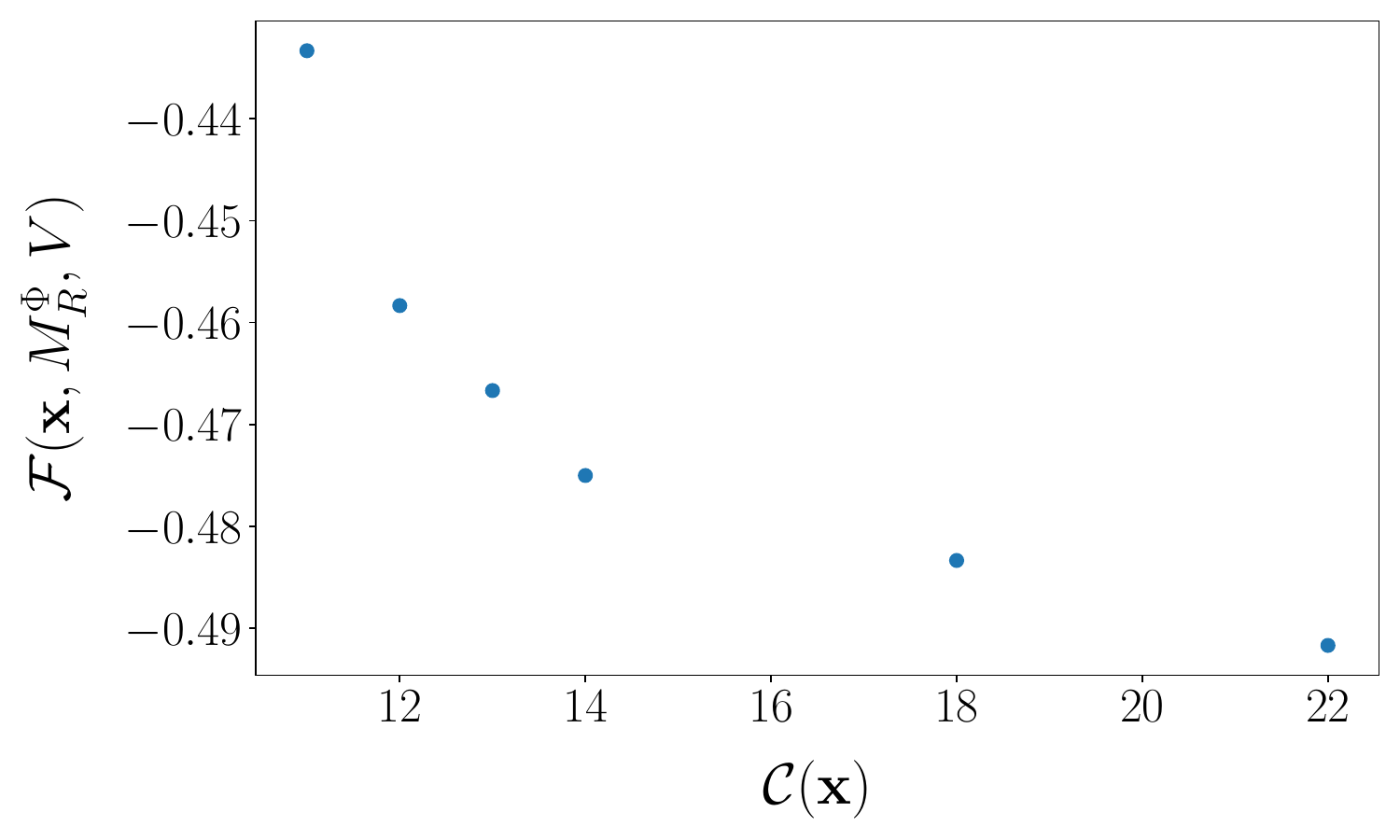} & \includegraphics[width=1\textwidth]{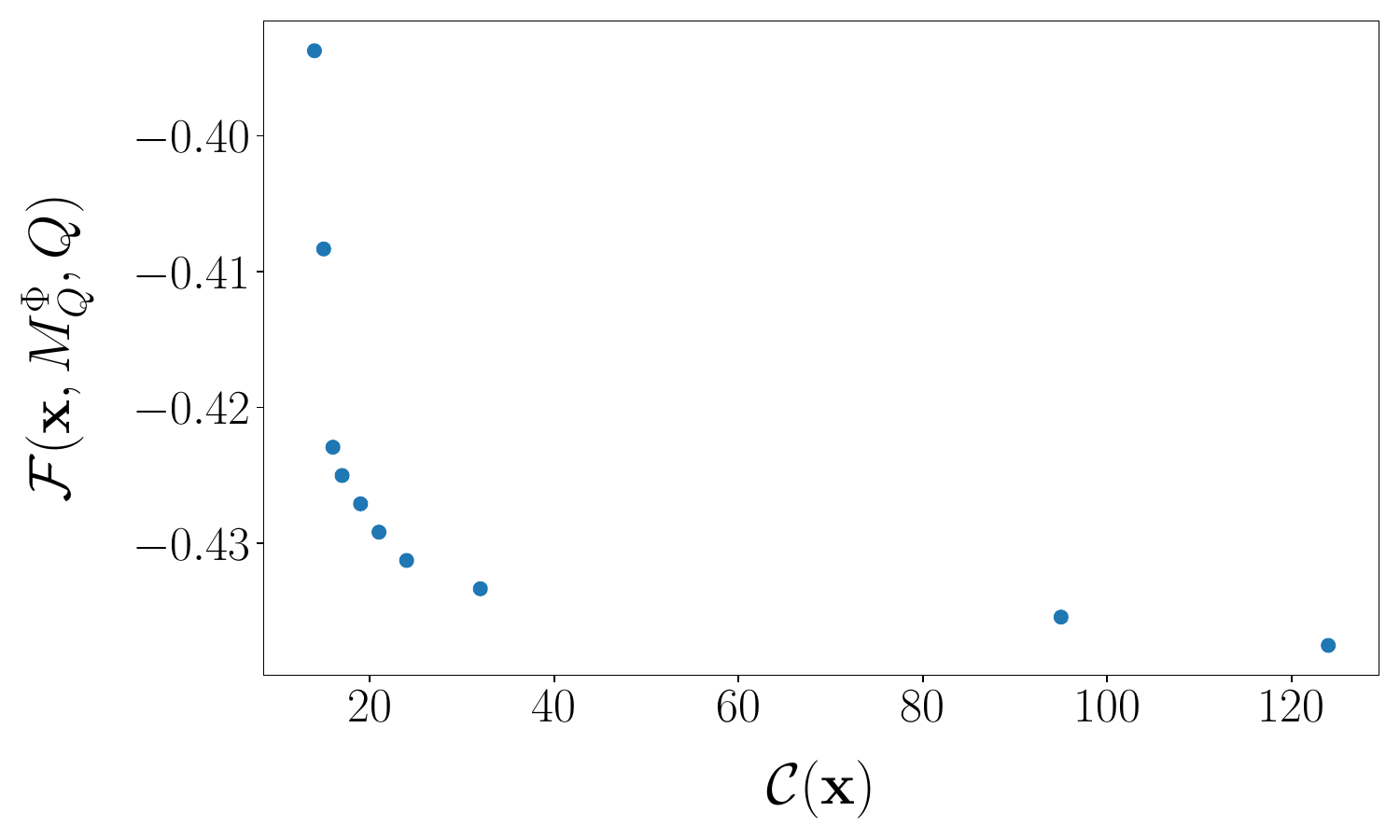}\\
			{\LARGE (a) PSEFS-MOEA-V1} & {\LARGE (b) PSEFS-MOEA-V2} \\
		\end{tabular}
	}
	\caption{Hipervolume evolution and Pareto front of PSEFS-MOEA-V1 (a) and PSEFS-MOEA-V2 (b) for the isolet dataset.}
	\label{fig:isolet}
\end{table*}

\begin{table*}[h]
	\centering
	\resizebox{\textwidth}{!}{
		\begin{tabular}{cc}
			
			\includegraphics[width=1\textwidth]{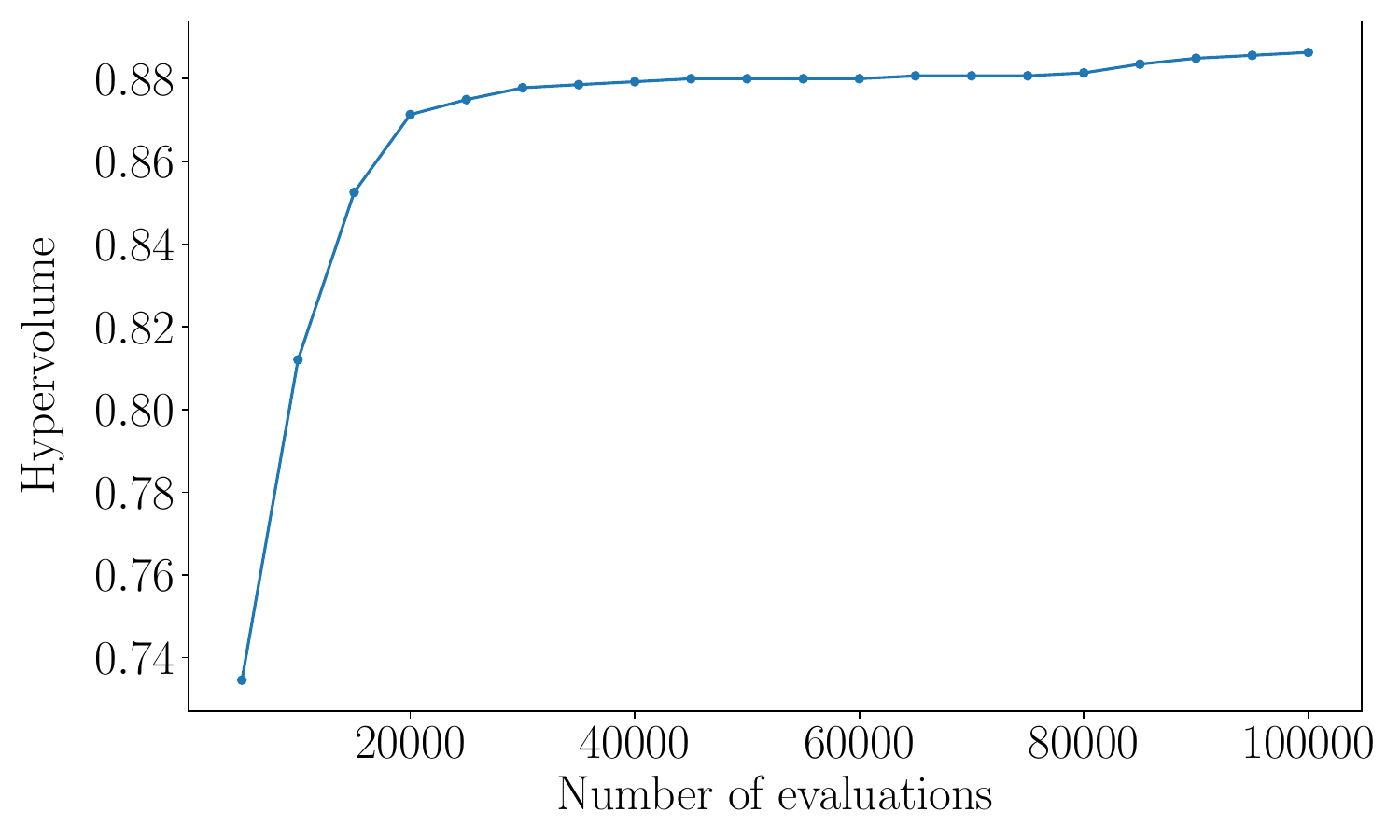} & \includegraphics[width=1\textwidth]{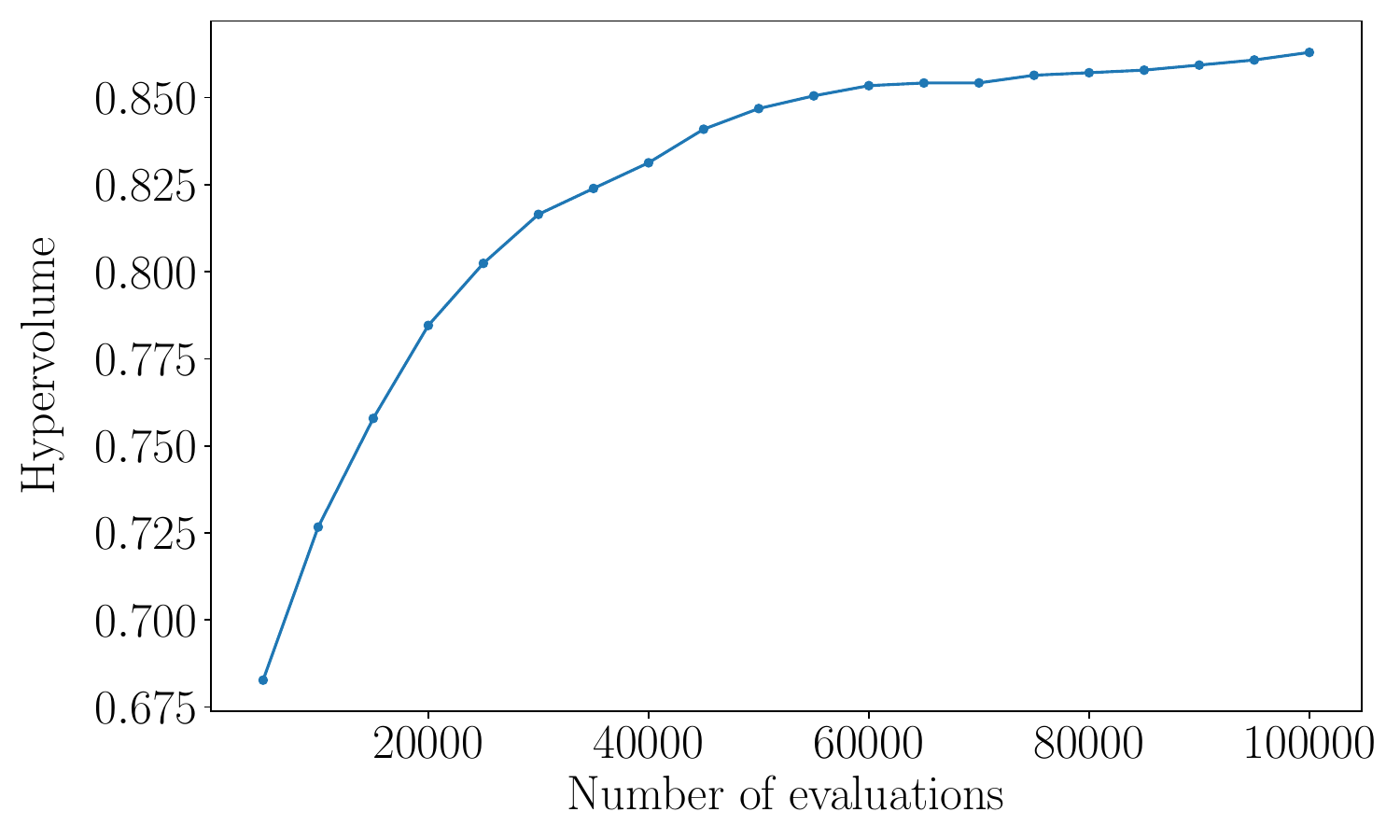}\\
			\includegraphics[width=1\textwidth]{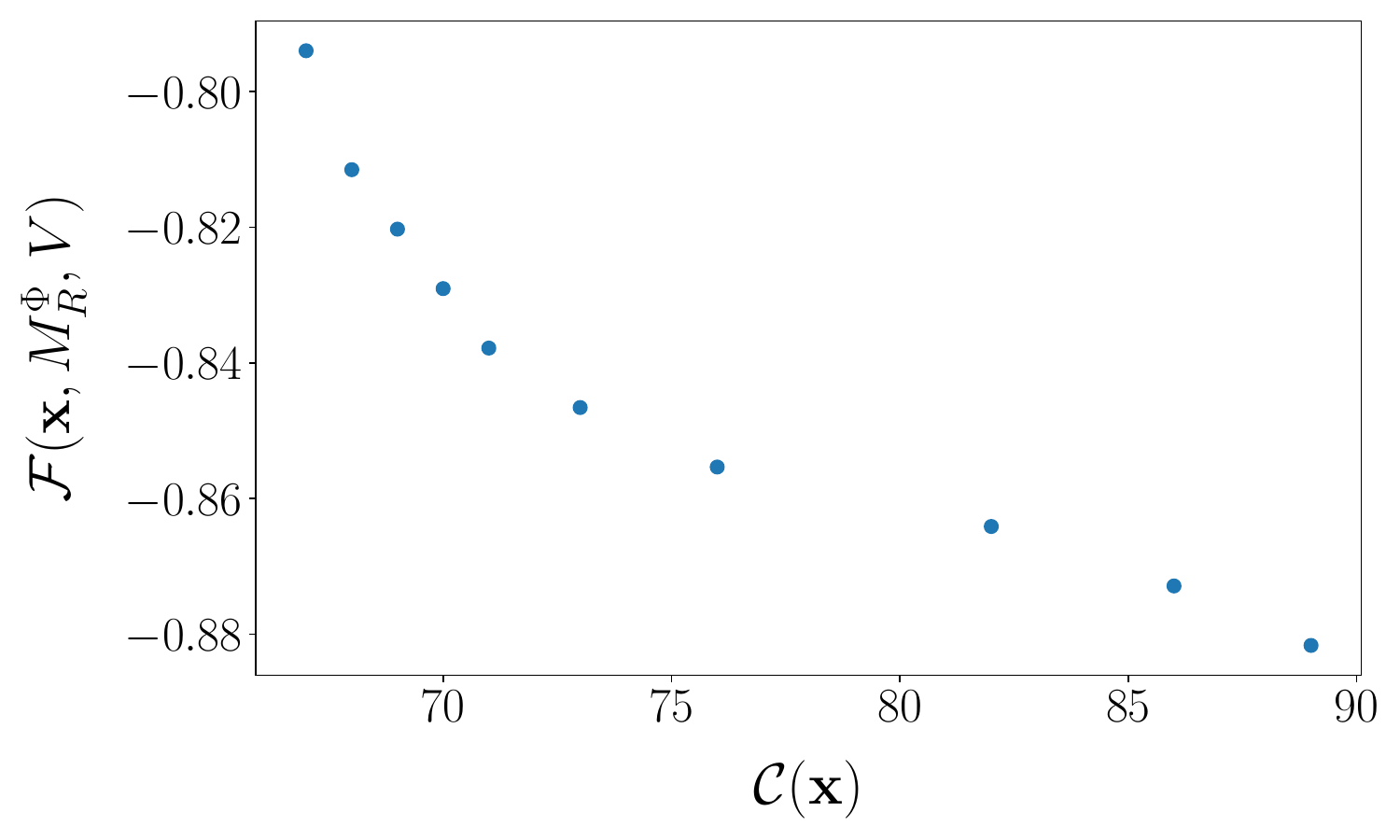} & \includegraphics[width=1\textwidth]{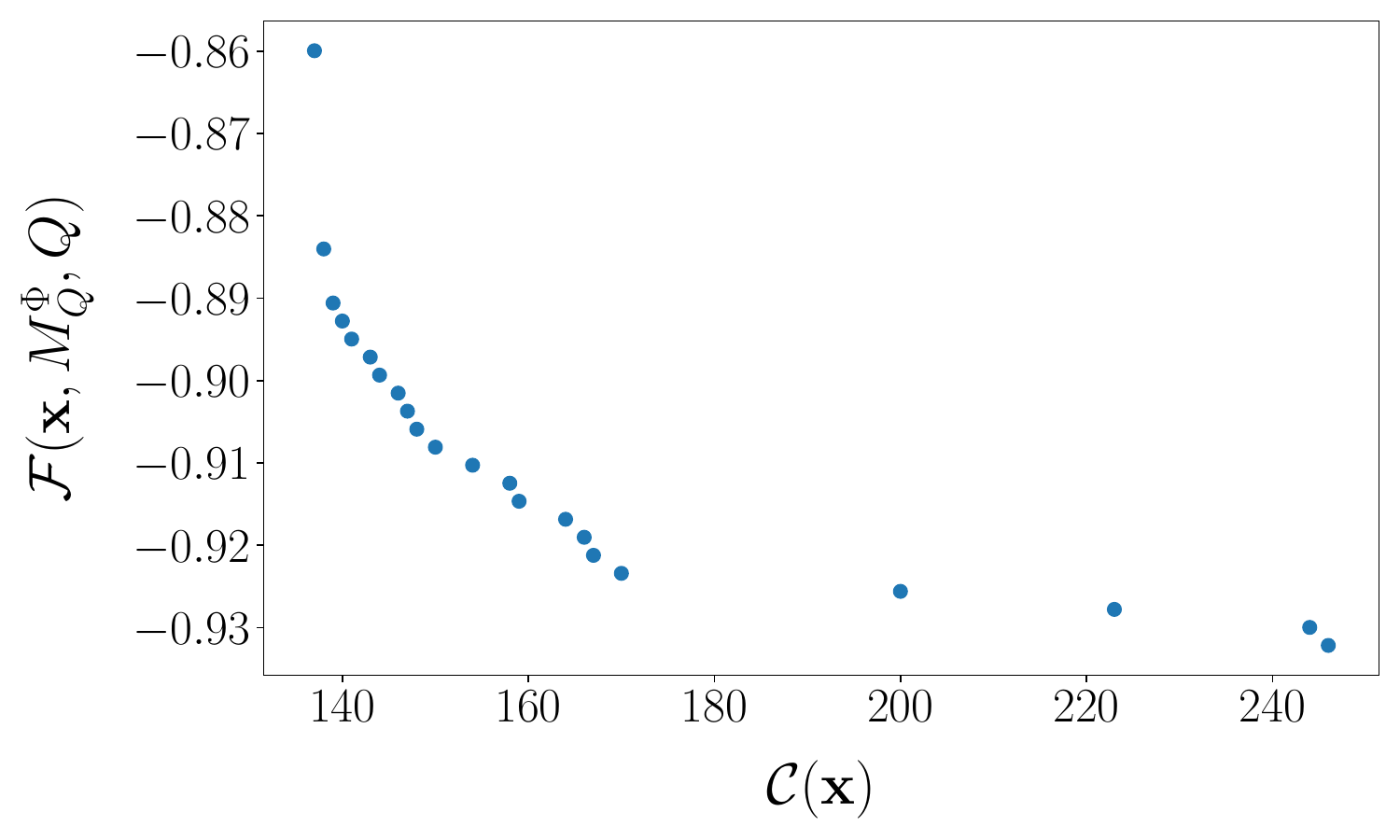}\\
			{\LARGE (a) PSEFS-MOEA-V1} & {\LARGE (b) PSEFS-MOEA-V2} \\
		\end{tabular}
	}
	\caption{Hipervolume evolution and Pareto front of PSEFS-MOEA-V1 (a) and PSEFS-MOEA-V2 (b) for the micro-mass dataset.}
	\label{fig:micro-mass}
\end{table*}

\begin{table*}[h]
	\centering
	\resizebox{\textwidth}{!}{
		\begin{tabular}{cc}
			
			\includegraphics[width=1\textwidth]{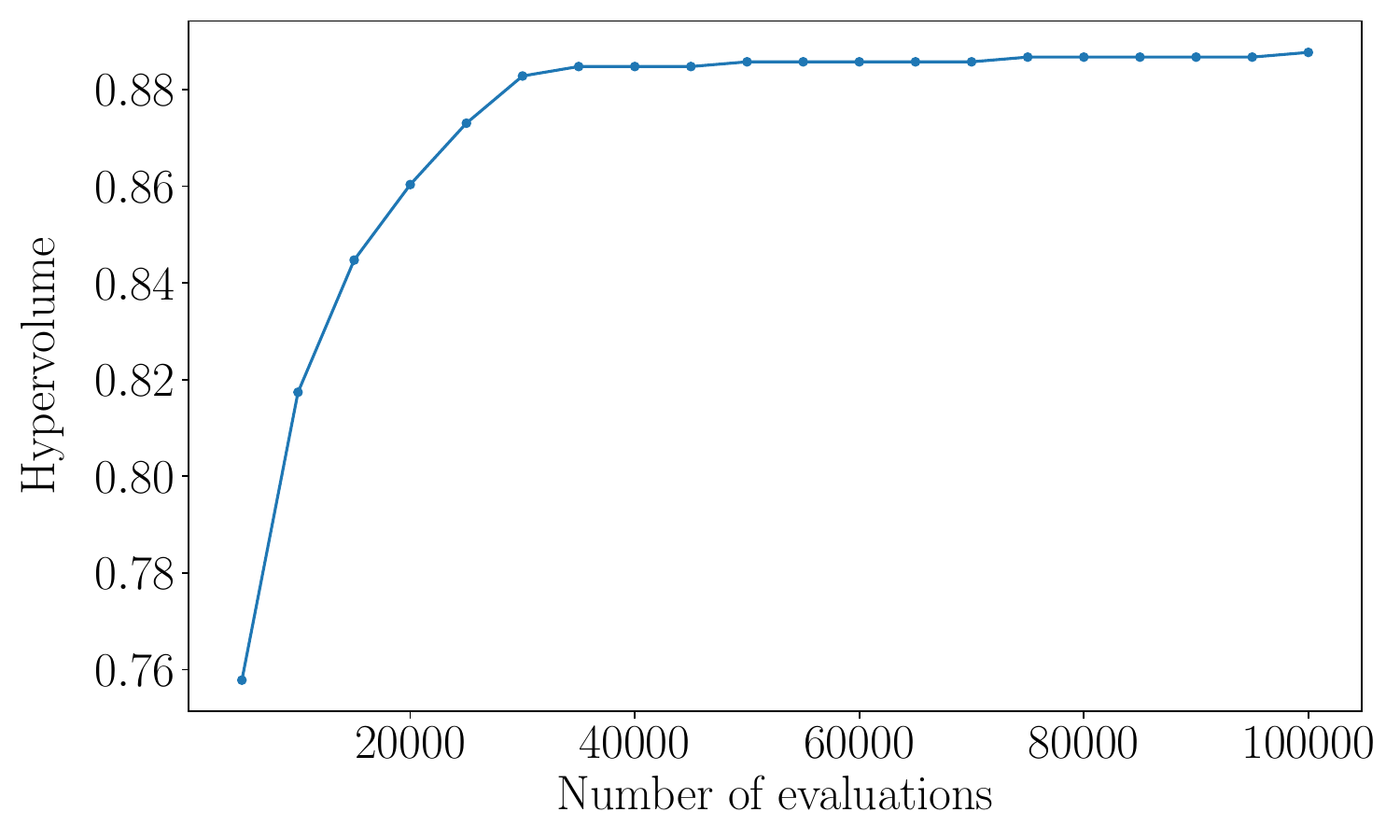} & \includegraphics[width=1\textwidth]{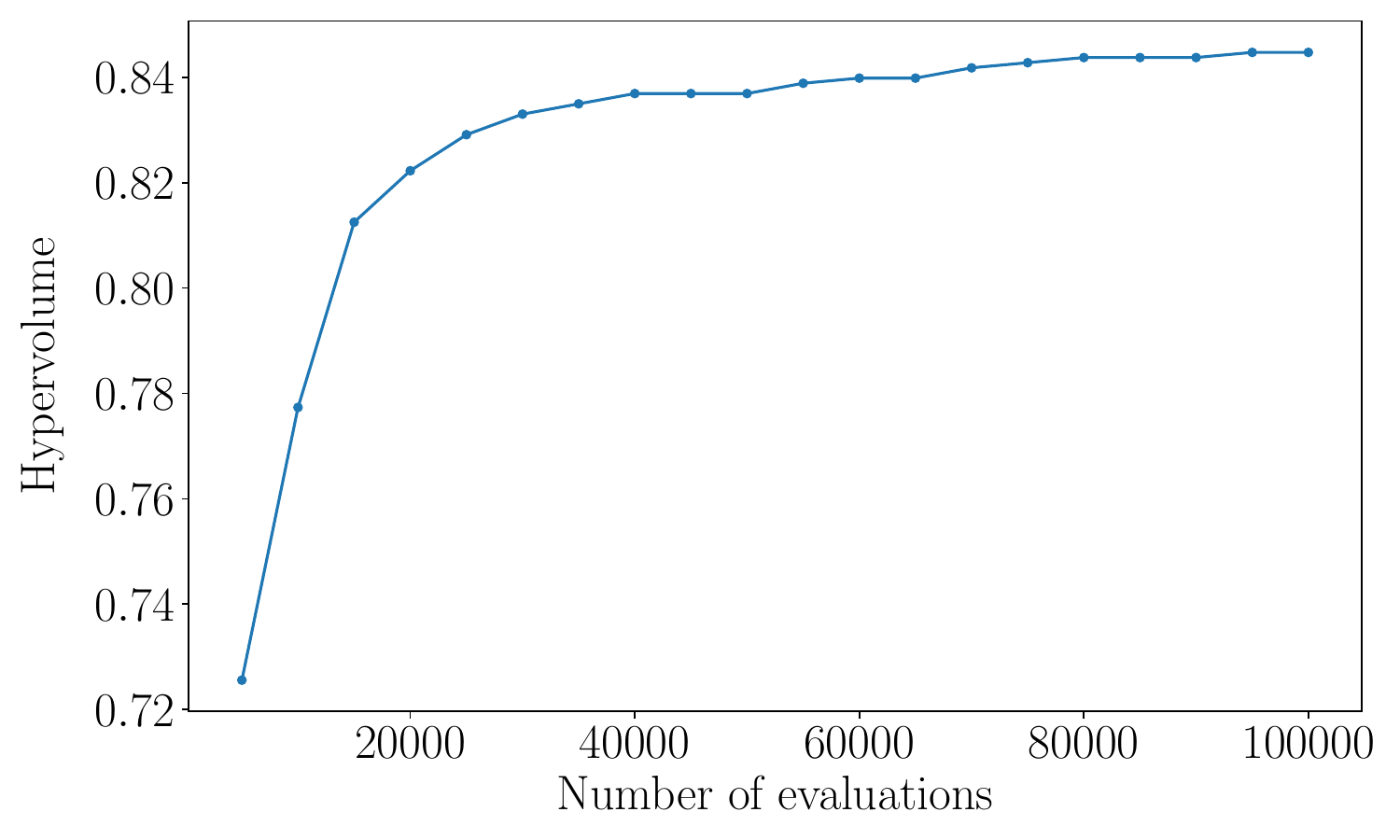}\\
			\includegraphics[width=1\textwidth]{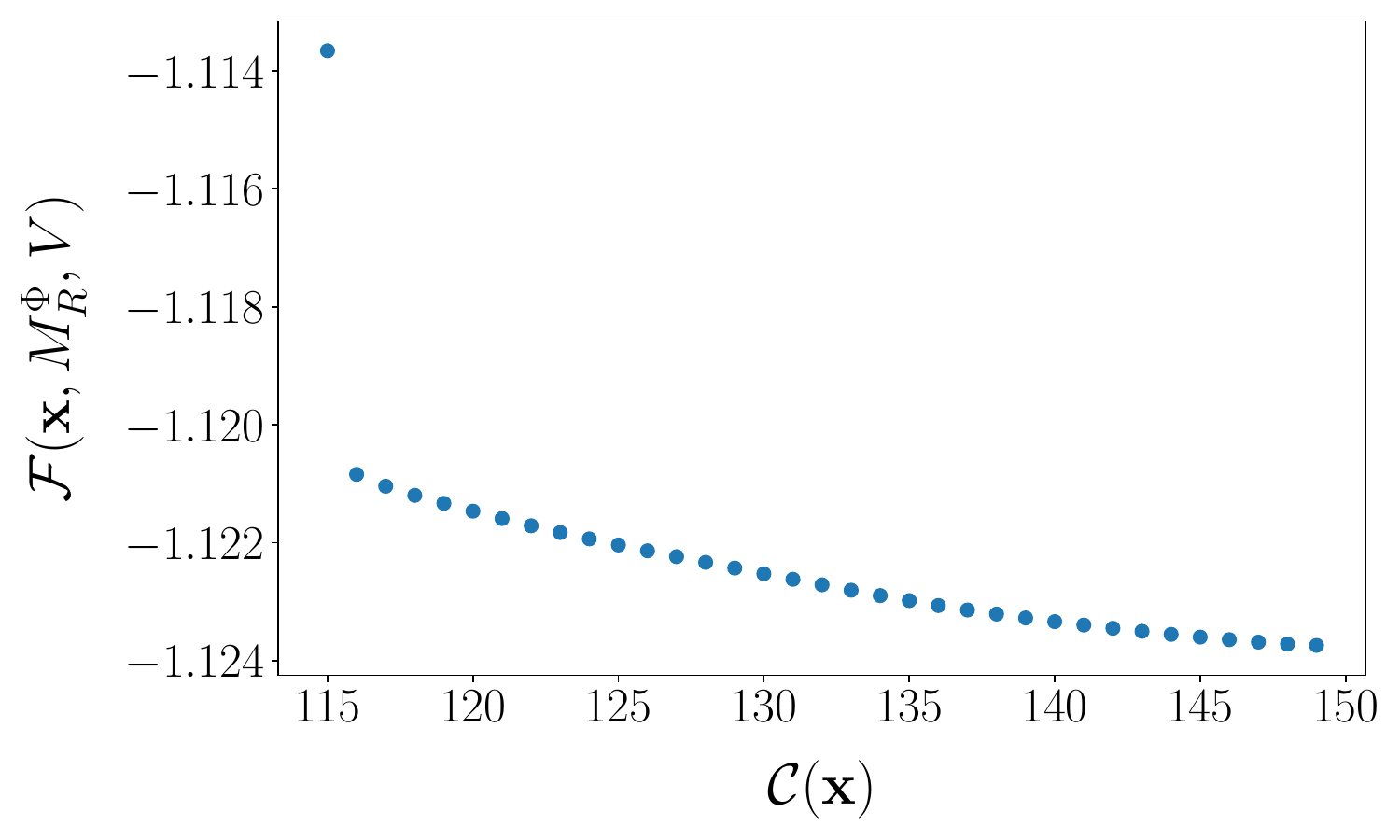} & \includegraphics[width=1\textwidth]{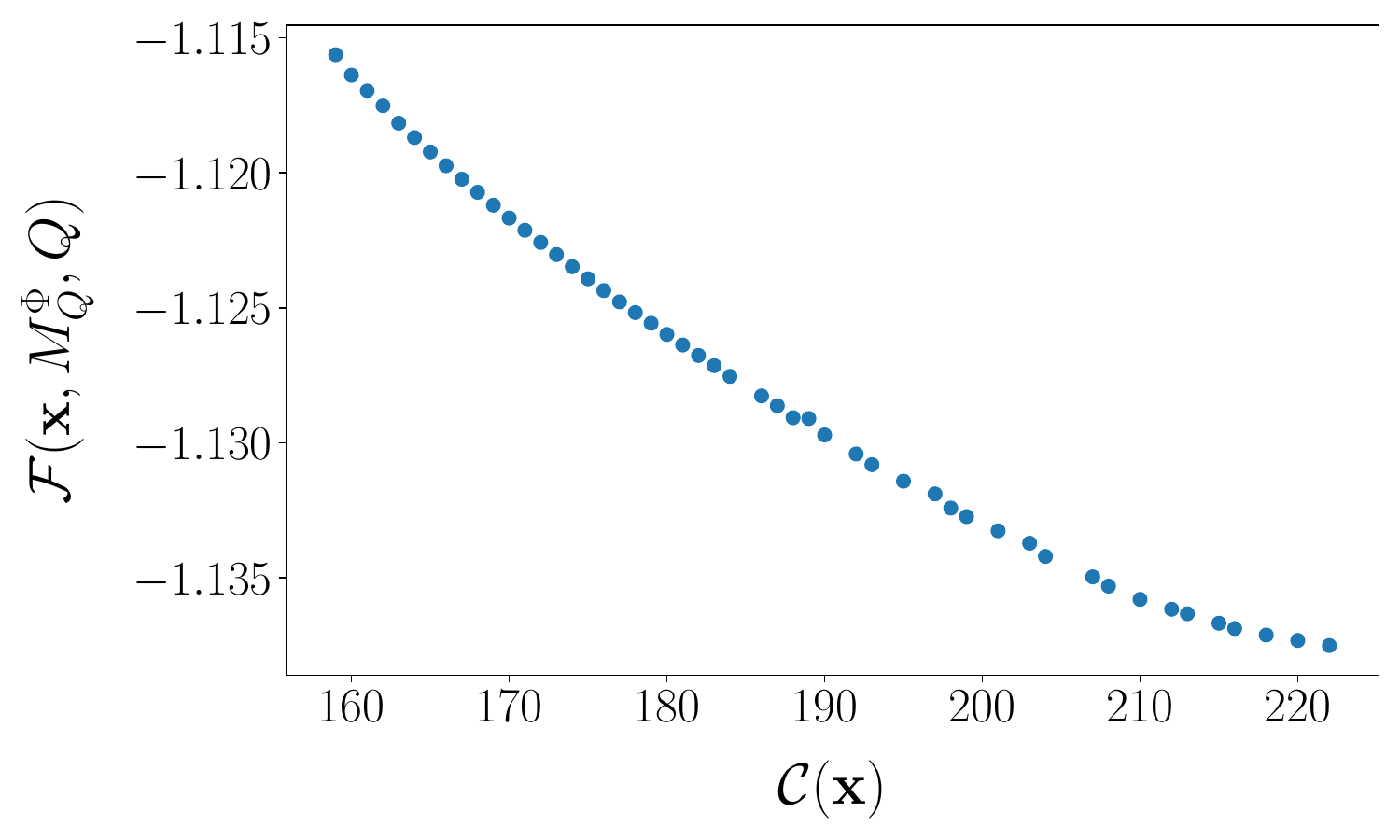}\\
			{\LARGE (a) PSEFS-MOEA-V1} & {\LARGE (b) PSEFS-MOEA-V2} \\
		\end{tabular}
	}
	\caption{Hipervolume evolution and Pareto front of PSEFS-MOEA-V1 (a) and PSEFS-MOEA-V2 (b) for the QSAR-TID-11054 dataset.}
	\label{fig:QSAR-TID-11054}
\end{table*}

In summary, the robustness of PSEFS-MOEA, both in classification and regression tasks, underscores its versatility and reliability as a FS method. Its strong performance across diverse problem types--ranging from imbalanced classification datasets to complex synthetic regression problems--makes it a highly effective approach for tackling real-world challenges characterized by diverse data distributions and varying levels of complexity.

\section{Conclusions and future work}  
\label{sec:conclusions}

In this study, we have proposed PSEFS-MOEA, a novel permutation-based feature selection  method. Unlike the conventional permutation feature importance, which evaluates individual attributes, PSEFS-MOEA assesses subsets of attributes using a multi-objective approach and employs a multi-objective evolutionary algorithm  as the search strategy for identifying optimal subsets. This innovative framework enables the exploration of more complex attribute interactions and dependencies, addressing limitations inherent to single-feature evaluation methods. The key contributions of the paper and conclusions are as follows

\begin{enumerate}

\item \textit{Two versions of PSEFS-MOEA}: We introduced two versions of the proposed method. PSEFS-MOEA-V1 evaluates the importance of attribute subsets on a validation set.
PSEFS-MOEA-V2 evaluates the importance directly on the training set.
The experimental results show that PSEFS-MOEA-V2 outperforms in classification tasks, while PSEFS-MOEA-V1 excels in regression tasks, highlighting the adaptability of the proposed framework to different problem types.

\item \textit{Comprehensive benchmarking}: The method was tested on a diverse set of 24 datasets, covering classification (binary, multi-class, and imbalanced problems) and regression tasks with increasing complexity. This diversity ensures a robust evaluation of PSEFS-MOEA across a wide range of real-world scenarios.

\item \textit{Superior performance}: Both versions of PSEFS-MOEA consistently outperformed nine well-established FS methods tailored for high-dimensional data. These baseline methods included filter approaches for both subset-based and single-feature evaluations, such as the conventional PFI. PSEFS-MOEA effectively reduced the number of features while improving model performance across metrics including ACC, BA, RMSE, and $R^2$.

\item \textit{Statistical validation}: Non-parametric statistical tests confirm the superiority of PSEFS-MOEA over competing methods in both classification and regression tasks.

\item \textit{Generalization capability}: The proposed method demonstrates strong generalization capabilities, as evidenced by low levels of overfitting. For classification tasks, PSEFS-MOEA showed significantly lower overfitting compared to all other FS methods. For regression tasks, overfitting levels were comparable to the best-performing methods.

\item \textit{Execution time}: The main drawback of PSEFS-MOEA is its longer execution time compared to other FS methods. This is an inherent limitation of its meta-heuristic, multi-objective, and population-based nature. However, the improved feature selection and model performance justify the computational expense in scenarios where accuracy and robustness are critical.
\end{enumerate}

To further enhance the capabilities of PSEFS-MOEA, the following research directions are proposed:

\begin{enumerate}
	\item \textit{Incorporating class weight optimization}: Develop an MOEA framework that integrates permutation-based feature selection with class weight optimization for addressing imbalanced classification problems more effectively.

	\item \textit{Ensemble learning integration}: Combine PSEFS-MOEA with ensemble learning strategies, such as voting or stacking, to enhance predictive performance and robustness.

	\item \textit{Scalability improvements}: Investigate strategies to reduce execution time without compromising the quality of feature selection, such as parallelization or hybrid approaches combining meta-heuristic and deterministic methods.
\end{enumerate}

By addressing these directions, PSEFS-MOEA can evolve into a more comprehensive and efficient tool, suitable for an even broader range of high-dimensional and complex datasets. The promising results achieved in this study validate the potential of permutation-based multi-objective FS methods and pave the way for future advancements in this domain.

\section*{Acknowledgements}
This paper is funded by the CALM-COVID19 project (Ref:
PID2022-136306OB-I00), grant funded by Spanish Ministry of
Science and Innovation and the Spanish Agency for Research.






\bibliographystyle{elsarticle-num}
\bibliography{references}

\appendix

\section{Abbreviations}

\begin{table}[h]
	\centering
	\begin{tabular}{ll}
			\hline
			\textbf{Abbreviation} & \textbf{Meaning}                                                   \\ 
			\hline
			ACC & Accuracy \\
			BA & Balanced Accuracy \\
			{FR}	  & {Feature Ranking} \\
			FS					  & Feature Selection \\
			MOEA                  & Multi-Objective Evolutionary Algorithm                             \\
			PFI & Permutation Feature Importance \\
			RF                    & Random Forest                                                      \\
			RMSE                  & Root Mean Square Error \\
			nRMSE & Normalized RMSE \\
			\hline
		\end{tabular}
\caption{Abbreviations.}
\end{table}

\end{document}